\documentclass{article}

\usepackage[main, final]{neurips_2026}

\usepackage[utf8]{inputenc} 
\usepackage[T1]{fontenc}    
\usepackage{hyperref}       
\usepackage{url}            
\usepackage{booktabs}       
\usepackage{amsfonts}       
\usepackage{nicefrac}       
\usepackage{microtype}      
\usepackage[table]{xcolor}  
\usepackage{multirow}
\usepackage{xspace}
\usepackage{amsmath}
\usepackage{graphicx}
\usepackage{wrapfig}
\usepackage{makecell}
\usepackage{pifont}
\usepackage{caption}
\usepackage{subcaption}

\usepackage{array}
\usepackage{tikz}             
\usepackage{pgfplots}         
\usepackage{pgf-pie}          
\usepackage{listings}        
\usepackage{ragged2e}        
\usepackage{comment}          
       
\usepackage{fontawesome}      
\usepackage{enumitem}        
\usepackage{titletoc}         
\usepackage{minitoc}  

\definecolor{lightblue}{RGB}{200, 230, 255}  
\definecolor{headerblue}{RGB}{150, 200, 255}
\newcolumntype{x}[1]{>{\centering\arraybackslash}p{#1pt}}
\newcolumntype{y}[1]{>{\raggedright\arraybackslash}p{#1pt}}
\newcolumntype{z}[1]{>{\raggedleft\arraybackslash}p{#1pt}}

\newcommand{\eg}{e.g.,\xspace}

\newcommand{\modelname}{{\textsc{SetCon}}\xspace}
\newcommand{\mjf}{$\mathcal{J}\&\mathcal{F}$\xspace}
\newcommand{\score}[1]{#1\phantom{$^\dagger$}}
\newcommand{\scoredag}[1]{#1$^\dagger$}

\makeatletter
\newcommand{\blfootnote}[1]{%
  \begingroup
  \renewcommand\thefootnote{}%
  \footnotetext{\hspace{-1.8em}#1}%
  \endgroup
}
\makeatother
\title{
SetCon: Towards Open-Ended Referring Segmentation via Set-Level Concept Prediction
}

\author{
Zhixiong Zhang$^{1,2,*}$ \quad
Yizhuo Li$^{1,2,*}$ \quad
Shuangrui Ding$^{3,\ddagger}$ \quad
\textbf{Yuhang Zang}$^{4,\dagger}$ \quad \\
\textbf{Shengyuan Ding}$^{5,2}$ \quad 
\textbf{Long Xing}$^{6}$ \quad
\textbf{Yibin Wang}$^{5,2}$ \quad
\textbf{Qiaosheng Zhang}$^{2,4}$ \quad
\textbf{Jiaqi Wang}$^{2,7,\dagger}$ \vspace{0.1cm} \\
$^1$Shanghai Jiao Tong University \quad 
$^2$Shanghai Innovation Institute \quad \\
$^3$The Chinese University of Hong Kong \quad 
$^4$Shanghai Artificial Intelligence Laboratory \quad \\
$^5$Fudan University \quad
$^6$University of Science and Technology of China \quad
$^7$JD.com \\
}
\begin{document}

\maketitle
\blfootnote{
$^*$ Equal Contribution \quad
$^\ddagger$ Project Lead \quad
$^\dagger$ Corresponding Author
}

\begin{abstract}
Referring segmentation grounds natural-language queries to pixel-level masks, but extending it to complex scenarios with multiple instances, cross-category groups, or open-ended target sets remains challenging. Previous Large Vision Language Model (LVLM)-based methods represent referred targets with one or more special tokens sequentially, treating multiple targets as separate outputs rather than a coherent set and offering little incentive to capture set-level properties such as completeness and mutual exclusivity. We \textit{reformulate} open-ended referring segmentation as explicit set-level concept prediction and propose \textbf{Set}-\textbf{Con}cept Segmentation (\textbf{\modelname}), which uses LVLM-generated natural-language concepts, instead of segmentation-specific tokens, as semantic conditions for joint mask-set decoding. A hierarchical semantic decomposition first predicts a shared set-level concept defining the target scope and then refines it into fine-grained concept groups aligned with target subsets. To support this, a two-stage annotation pipeline augments existing reasoning segmentation datasets with hierarchical semantic supervision (236k samples, 784k concept phrases). \modelname achieves state-of-the-art results on image benchmarks (\textbf{+3.3 gIoU} on gRefCOCO, \textbf{+12.1 gIoU} on MUSE), with margins that grow as the number of referred targets increases. The concept interface also transfers to video under a detect-and-track setting, yielding new state-of-the-art results on seven referring video benchmarks, including \textbf{+10.9 \mjf} on MeViS and \textbf{+12.4 \mjf} on Ref-SeCVOS. The code, model checkpoints, and dataset annotations are open-sourced \href{https://github.com/rookiexiong7/SetCon}{here}.
\end{abstract}

\section{Introduction}

Referring segmentation~\cite{kazemzadeh2014referitgame,mao2016generation,liu2023gres,khoreva2018video,seo2020urvos,ding2023mevis} aims to predict pixel-level masks for the targets described by a natural-language query, coupling language understanding with dense visual grounding. This fine-grained capability underpins applications such as interactive image and video editing~\cite{li2023gligen,tu2025videoanydoor}, industrial inspection~\cite{gu2024anomalygpt}, AR/VR interaction~\cite{xiu2025viddar}, and embodied perception~\cite{driess2023palm,zitkovich2023rt}. Building on large vision-language models (LVLMs), a growing line of work~\cite{lai2024lisa,rasheed2024glamm,yuan2025sa2va,wang2025xsam} extends referring segmentation beyond single-expression grounding toward more \textit{reasoning-intensive} settings. A representative paradigm~\cite{lai2024lisa,rasheed2024glamm,yuan2025sa2va,wang2025xsam,xia2024gsva,ren2024pixellm} appends a special $\texttt{[SEG]}$ token to the LVLM, whose hidden state is decoded into a mask, providing an \textbf{implicit} interface between language reasoning and pixel prediction. This design yields strong results on standard referring and reasoning segmentation benchmarks~\cite{kazemzadeh2014referitgame,mao2016generation,lai2024lisa}, which are nevertheless dominated by \textit{single-target} queries.

\begin{figure}[t]
    \centering
    \includegraphics[width=\textwidth]{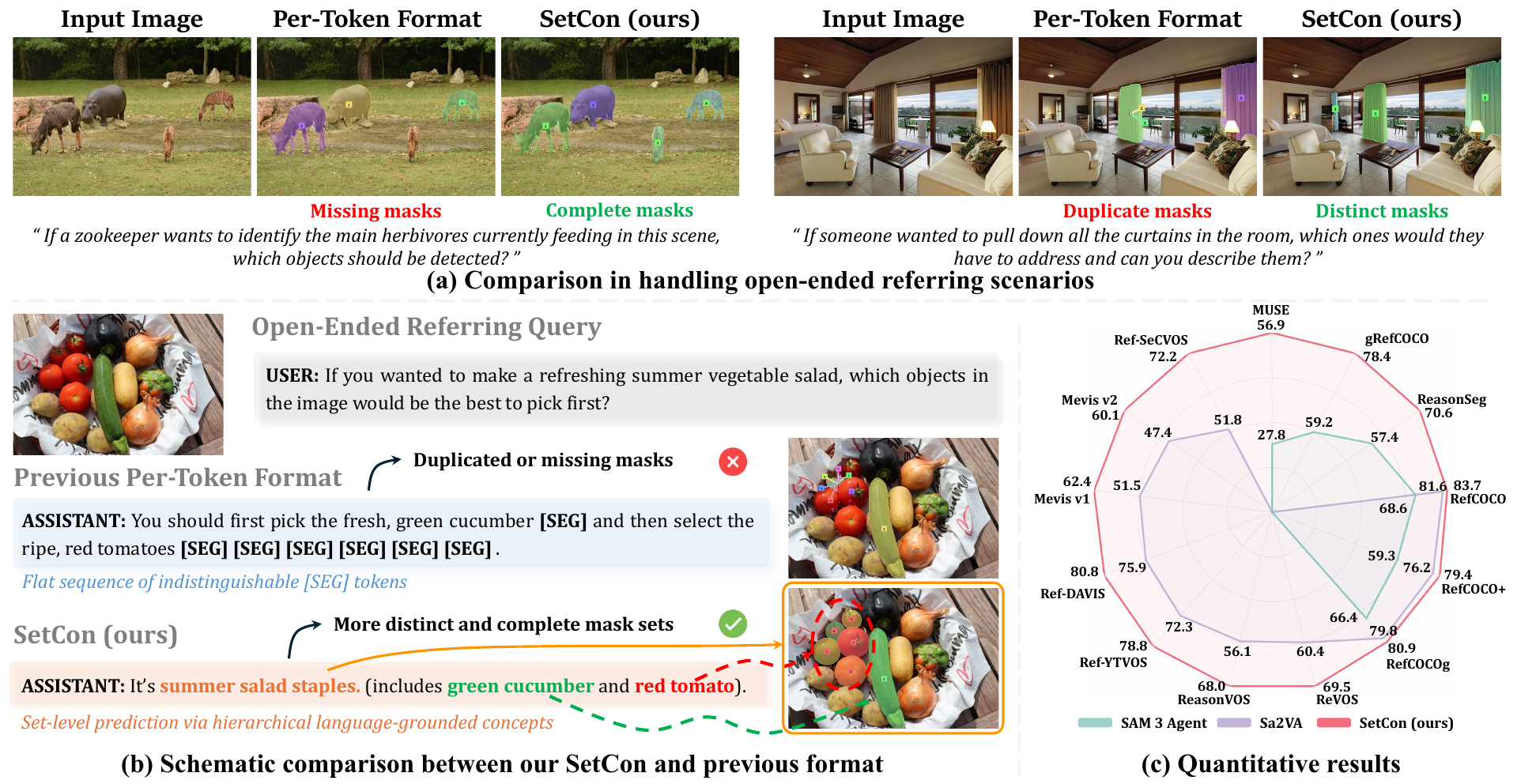}
    \vspace{-12pt}
    \caption{Overview of \textbf{Set}-\textbf{Con}cept Segmentation (\textbf{\modelname}). 
    \textbf{(a)} Compared to previous per-token formulation, \modelname produces more distinct and complete mask sets in open-ended scenarios.
    \textbf{(b)} \modelname predicts interpretable, hierarchical set-level concepts and uses them as semantic conditions for mask-set decoding instead of indistinguishable special tokens.
    \textbf{(c)} Quantitative results show that \modelname achieves the best performance on image and video referring segmentation benchmarks. 
    }
    \label{fig:teaser}
    \vspace{-12pt}
\end{figure}

However, real-world queries are often \textit{open-ended}, requiring reasoning over a heterogeneous set of plausible targets across multiple categories (e.g., picking ingredients for a salad), rather than locating a single object. Extending the prevailing paradigm to such scenarios remains challenging.
Existing methods~\cite{xia2024gsva,ren2024pixellm,lai2024lisa,rasheed2024glamm} typically represent referred targets with one $\texttt{[SEG]}$ token per target and decode the corresponding masks independently. This \textbf{per-token} formulation treats targets as separate outputs rather than a coherent semantic set, offering little incentive to capture set-level properties such as \textit{completeness}, \textit{cardinality}, and \textit{mutual exclusivity}, and may yield duplicated or missing masks and unstable target-to-mask assignment (Fig.~\ref{fig:teaser}(a)). To probe this limitation, we conduct a pilot study by extending a representative LVLM-based segmentation model, Sa2VA~\cite{yuan2025sa2va}, to the multi-target setting on the MUSE benchmark~\cite{ren2024pixellm}. We observe that performance degrades sharply as the number of referred targets grows, and that the projected $\texttt{[SEG]}$ embeddings cluster predominantly by \textit{spatial position} rather than \textit{semantic category}, suggesting that the implicit token interface encodes spatial layout more than the semantic structure of the referred set (see Sec.~\ref{sec:preliminary} for the detailed analysis).

Motivated by these findings, we reformulate open-ended referring segmentation as \textit{explicit set-level concept prediction}. We present \textbf{Se}t-\textbf{Con}cept \textbf{S}egmentation (\textbf{\modelname}), an end-to-end framework that couples LVLM-side language reasoning with set-level mask decoding through a \textbf{language-grounded concept interface} (Fig.~\ref{fig:teaser} (b)). Given a language query and an image, \modelname produces a textual response in which the referred targets are surfaced as explicit natural-language concepts; the hidden states of these concept spans are then projected into the segmentation decoder as semantic conditions, and the full target mask set is decoded jointly. By replacing segmentation-specific tokens learned during finetuning with concepts drawn from the LVLM's own vocabulary, \modelname grounds every mask in an interpretable semantic anchor and reuses the model's \textit{pretrained semantic geometry}, rather than relearning it from a newly initialized token. To handle cross-category scenarios, the concepts are organized via a \textbf{hierarchical semantic decomposition}: \modelname first emits a \textit{set-level concept} that summarizes the overall target scope, and then decomposes it into \textit{fine-grained sub-concepts}, one per target subset. This \textbf{coarse-to-fine} structure delineates the boundary of the target set while separating its sub-categories, improving both coverage and inter-category discrimination.

\modelname extends naturally to \textbf{referring video segmentation} under a detect-and-track formulation~\cite{ravi2025sam2,carion2025sam}: the LVLM is invoked once per video to produce motion-aware concepts, which are shared across frames as fixed semantic conditions for per-frame decoding and memory-based mask propagation. This design keeps temporal inference efficient while preserving identity consistency across frames.

Training the concept interface requires linguistic supervision at both the set and the sub-category level, which existing reasoning segmentation datasets~\cite{lai2024lisa,ren2024pixellm} do not provide: their annotations are instance-level and rely on closed-vocabulary labels. To bridge this gap, we build \textit{hierarchical semantic supervision} via a two-stage annotation pipeline: every target is annotated with a free-form \textit{sub-category phrase} grounded in its instance masks and the original query, and these phrases are organized under a global \textit{set-level concept} that summarizes the target scope. Candidate annotations are further cleaned by rule-based filtering and multiple rounds of manual spot-checking. The resulting corpus has $236{,}396$ samples and $784{,}809$ concept phrases, with over $80\%$ of the samples involving more than one sub-category, providing the set-level signal that prior corpora lack and supplying explicit concept-to-mask correspondence for open-ended multi-target and cross-category grounding.

We evaluate \modelname on six image and seven video referring segmentation benchmarks (see Fig.~\ref{fig:teaser} (c)). On multi-object image segmentation, \modelname achieves SOTA performance on gRefCOCO and MUSE, surpassing the previous best by \textbf{+3.3 gIoU} on gRefCOCO \textit{val} and \textbf{+12.1 gIoU} on the challenging MUSE \textit{val}, with the margin widening as the number of referred targets grows. On standard single-target benchmarks (RefCOCO/+/g and ReasonSeg), \modelname remains competitive and attains the best average, indicating that our method does not regress on conventional settings. On video, \modelname achieves new SOTA results across all \textit{seven} referring video segmentation benchmarks, with substantial gains on the challenging MeViS (\textbf{+10.9 \mjf}) and Ref-SeCVOS (\textbf{+12.4 \mjf}) settings, where stable semantic anchors are most beneficial for long-horizon temporal association.

Our contributions are:
\textbf{1)} Through a pilot study on a representative LVLM-based segmentation model, we identify two limitations of the prevailing $\texttt{[SEG]}$-token paradigm in open-ended referring scenarios.
\textbf{2)} We reformulate open-ended referring segmentation as \textit{explicit set-level concept prediction} and propose \modelname, an end-to-end framework that couples LVLM-side reasoning with set-level mask decoding through a language-grounded concept interface, organized hierarchically into a set-level concept and fine-grained sub-concepts.
\textbf{3)} \modelname achieves SOTA performance on six image and seven video referring segmentation benchmarks, with the largest gains on multi-target and cross-category settings, while remaining competitive on conventional single-target benchmarks.

\begin{figure}[t]
    \centering
    \includegraphics[width=\textwidth]{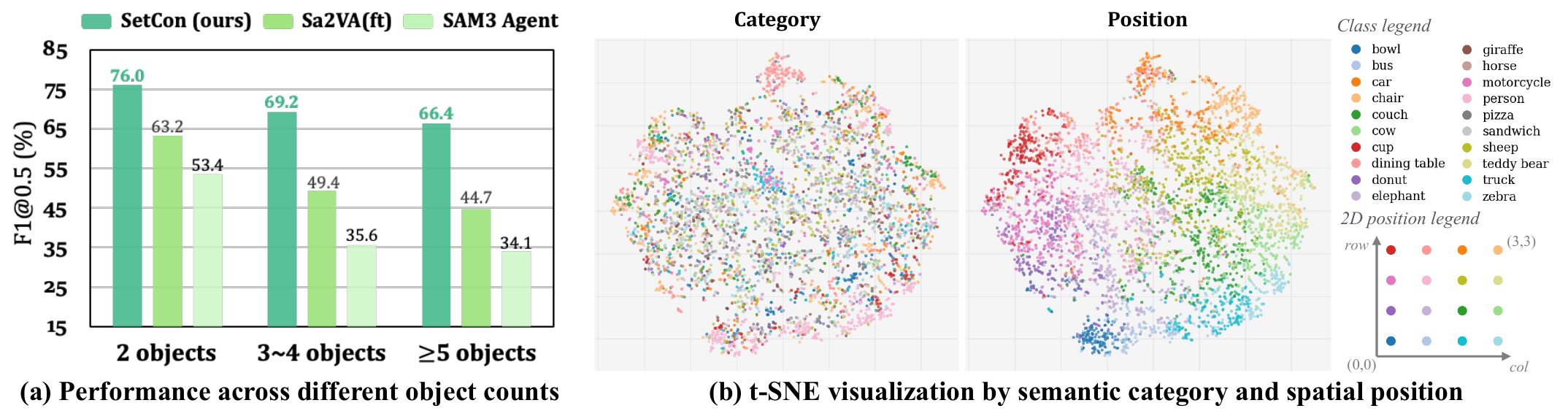}
    \caption{\textbf{Pilot study motivating explicit set-level concept prediction.} \textbf{(a)} special-token baselines degrade sharply as the target count grows, while \modelname remains comparatively stable. \textbf{(b)} t-SNE projection of $\texttt{[SEG]}$ representations from the Sa2VA-based baseline, colored by semantic category and 2D spatial position; clusters align more clearly with position than with category.}
    \label{fig:preliminary}
    \vspace{-12pt}
\end{figure}
\section{Related Work}

\noindent \textbf{Open-vocabulary Grounding.}
Open-vocabulary grounding aims to localize visual entities specified by natural-language concepts, and has been advanced by vision-language pretraining, text-conditioned detection, and region-text alignment~\citep{kamath2021mdetr,zhong2022regionclip,li2022grounded,minderer2022simple,liu2024grounding}. Subsequent segmentation models extend this capability from boxes or regions to open-vocabulary mask prediction with text or visual prompts~\citep{luddecke2022image,liang2023open,zou2023segment,zou2023generalized,wu2024general}. Recent promptable foundation models further broaden this paradigm across image and video domains, enabling promptable and concept-level segmentation~\citep{kirillov2023segment,ravi2025sam2,ding2025sam2long,carion2025sam}. However, these methods typically treat each prompt or concept separately, leaving open-ended cross-category scenarios underexplored. In contrast, \modelname predicts hierarchical concepts as explicit semantic conditions to decompose open-ended target sets and guide joint mask-set decoding.

\noindent \textbf{Referring Segmentation.}
Referring segmentation aims to localize pixel-level masks for targets described by natural-language expressions~\citep{kazemzadeh2014referitgame,mao2016generation}. Early methods ground expressions through cross-modal fusion between visual features and linguistic queries~\citep{hu2016segmentation,liu2017recurrent,yu2018mattnet}, while later works improve language-aware dense prediction with stronger vision-language representations~\citep{yang2022lavt,ding2022vlt,zou2023segment}. Subsequently, some works extend the task to more generalized referring scenarios~\cite{liu2023gres,hu2023beyond} and part-level formulations~\cite{wang2024unveiling}. More recently, LVLM-based methods~\cite{lai2024lisa,chen2024sam4mllm,ren2024pixellm} further broaden the task toward reasoning segmentation with open-ended instructions. In the video domain, referring video object segmentation additionally requires maintaining temporal consistency across evolving frames~\cite{khoreva2018video,seo2020urvos,ding2023mevis,ding2022motion,qian2023semantics,ding2024betrayed,ding2023prune}. In this work, we provide a unified concept-based framework for open-ended image and video referring segmentation.

\noindent \textbf{LVLMs for Fine-grained Perception.}
Large vision-language models (LVLMs)~\cite{singh2025openai,team2024gemini,bai2025qwen3,abtahizadeh2026muse,ding2026arm,ding2026beyond,han2026lomo,xing2026caprl} have recently been applied to fine-grained perception to bridge semantic understanding and dense prediction. Pioneered by LISA~\cite{lai2024lisa}, many methods introduce a dedicated \texttt{[SEG]} token as an implicit interface for mask decoding, and further extend this design to grounded conversation~\cite{zhang2024omg,rasheed2024glamm}, visual understanding~\cite{yuan2025sa2va,wang2025xsam}, multi-target segmentation~\cite{ren2024pixellm,xia2024gsva,jang2025mmr}, and temporal scenarios~\cite{bai2024one,yan2024visa,zhang2025sec}. Recent works also explore more explicit reasoning forms, such as reformulating segmentation as text generation~\cite{lan2025text4seg} or producing intermediate reasoning steps and geometric prompts~\cite{lu2025rsvp,liu2025segzero,liu2026visionreasoner}. Instead of relying on special tokens or external geometric prompting, we adopt hierarchically organized natural-language concepts as semantic conditions for mask-set prediction, leveraging the pretrained semantic structure of LVLMs to better handle open-ended referring segmentation.

\section{Methodology}
\label{sec:method}

\begin{figure}[t]
    \centering
    \includegraphics[width=0.98\textwidth]{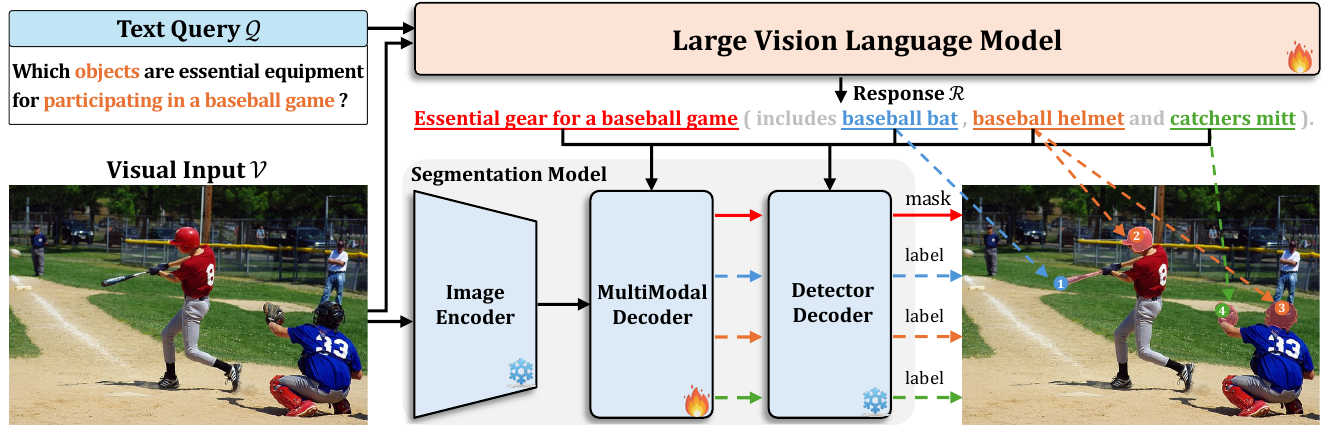}
    \caption{\textbf{Architecture of \modelname.} Given a text query $\mathcal{Q}$ and visual input $\mathcal{V}$, the LVLM produces a response $\mathcal{R}$ containing a global set-level concept and its decomposed sub-category concepts. The multimodal decoder is trainable while the image encoder and detector decoder remain frozen, to jointly predict the mask set with per-target labels.
    }
    \label{fig:method}
    \vspace{-12pt}
\end{figure}

\subsection{Preliminary Study}
\label{sec:preliminary}

Given a visual input $\mathcal{V}$ (an image or a video clip) and a natural-language query $\mathcal{Q}$, referring segmentation aims to predict the binary mask set $\mathcal{M}$ of all targets that satisfy $\mathcal{Q}$. A common practice in recent LVLM-based methods~\citep{lai2024lisa,ren2024pixellm,xia2024gsva,yuan2025sa2va,wang2025xsam} is to introduce a special $\texttt{[SEG]}$ token into the LVLM vocabulary and use it as an implicit interface for mask prediction. In this paradigm, the model generates a flat sequence of $\texttt{[SEG]}$ tokens for each referred target, and the hidden state $\mathbf{h}_{\texttt{seg}}$ of each token is separately fed into a mask decoder $\mathcal{D}$ to produce a mask $m=\mathcal{D}(\mathcal{V},\mathbf{h}_{\texttt{seg}})$.

However, it is highly counter-intuitive to encode the distinct semantic and spatial properties of multiple diverse targets into the hidden states of identical, repetitively generated $\texttt{[SEG]}$ tokens. Such a trivial design inevitably limits the representation capacity and risks feature ambiguity. To further investigate the limitation of the implicit $\texttt{[SEG]}$-token paradigm under open-ended scenarios, we conduct a preliminary study on a representative model Sa2VA~\citep{yuan2025sa2va}. Following~\cite{ren2024pixellm}, we extend it to the open-ended multi-target setting by sequentially predicting multiple special $\texttt{[SEG]}$ tokens within a single response. We finetune the model on the MUSE~\citep{ren2024pixellm} training set and evaluate it on the test set.

We categorize samples by the number of referred targets and report the F1 score at an IoU threshold of 0.5. As shown in Fig.~\ref{fig:preliminary}(a), performance degrades sharply as the target count grows, indicating the per-token formulation is ill-posed to more challenging multi-target segmentation settings, where the model has to reason about the target set holistically to ensure completeness and avoid duplicates. Besides, the absence of explicit semantic distinctions among repetitively generated tokens makes their alignment with ground-truth instances inherently ambiguous, which hinders model optimization.

To further investigate what these implicit tokens encode, we apply t-SNE to the projected $\texttt{[SEG]}$ representations on a deduplicated, class-balanced subset of RefCOCO. As illustrated in Fig.~\ref{fig:preliminary}(b), the embeddings reveal clearer cluster structure under spatial-position coloring than under semantic-category coloring, and spatially adjacent regions remain close in the embedding space. This indicates that the implicit $\texttt{[SEG]}$ representations are organized predominantly by spatial layout rather than by semantics. Consequently, although sufficient for mask decoding in few-target cases, they form a suboptimal interface for more complex, open-ended scenarios. 

\subsection{\modelname}
To overcome these inherent limitations, we propose a paradigm shift from implicit token generation to explicit semantic grounding, directly utilizing natural language as an interpretable and discriminative interface for complex multi-target segmentation.
We propose \modelname, an end-to-end framework that jointly grounds an open-ended target set with explicit, interpretable semantics. An overview of the architecture is shown in Fig.~\ref{fig:method}.

\noindent \textbf{Explicit Set-Level Concept Prediction.}
Instead of relying on implicit $\texttt{[SEG]}$ tokens as a latent signal for mask decoding, \modelname reads the target concepts directly from its own textual output. Specifically, the model first generates a response $\mathcal{R}$ in which each referred concept $\mathcal{C}$ is expressed as a free-form language phrase and delimited by special markers $\texttt{<ref>}\!\dots\!\texttt{</ref>}$. Each concept is bound to an entire \emph{set} of semantically coherent instances rather than to a single target, so that $\mathcal{R}$ explicitly encodes the granularity of set-level mask prediction. The hidden states $\mathbf{H}$ of the concept tokens are then projected into the decoder feature space as semantic conditions $\tilde{\mathbf{H}}$, and the segmentation module $\mathcal{D}$ produces the associated mask set $\mathcal{M} = \mathcal{D}(\mathcal{V}, \tilde{\mathbf{H}})$. Compared with implicit-token interfaces, this formulation not only attaches an interpretable semantic anchor to each predicted target set, but also reuses the LVLM's pretrained semantic representations rather than learning them from scratch with newly initialized tokens. Moreover, by expressing concepts in free-form language instead of a fixed taxonomy, it naturally supports open-ended, cross-category scenarios in which the target cardinality, identity, and semantic scope are dynamically determined by the query and visual context.

\noindent \textbf{Hierarchical Semantic Decomposition.}
Many real-world queries refer not to a single category but to a group spanning several sub-categories, e.g.\ ``all animals in the picture''. Enumerating every sub-category as a flat sequence of concepts tends to yield incomplete target sets. \modelname therefore organizes the predicted concepts hierarchically: it first emits a set-level concept that summarizes the shared semantics of the entire target group, and then decomposes it into a list of fine-grained sub-category concepts. Concretely, the first predicted concept serves as the global representation $\tilde{\mathbf{H}}_0$ for joint mask generation, while the remaining concepts $\{\tilde{\mathbf{H}}_i\}_{i=1}^{N}$ correspond to the $N$ semantic subsets used for label assignment, with $N$ freely determined by the LVLM rather than fixed in advance. During mask decoding, each sub-category representation is fused with $\tilde{\mathbf{H}}_0$ so that fine-grained predictions are conditioned on the shared semantic scope. This coarse-to-fine design is intended to improve both coverage and discrimination: the global concept delineates what belongs to the target set, while the sub-category concepts separate the subsets within it.

\noindent \textbf{Training and Inference.}
\modelname is trained end-to-end with an auto-regressive language-modeling loss on the response $\mathcal{R}$ and a DETR-style set-prediction loss~\citep{carion2020detr} on the mask predictions of each concept group. For video clips, we randomly sample sparse frames per clip and share a single concept response across them, as the referred semantics remain consistent over time. For no-target samples, we replace the ground-truth concepts with ``no target'' in $50\%$ of the cases and keep the original concepts in the rest, encouraging the LVLM and segmentation model $\mathcal{D}$ to learn abstention from complementary signals. At inference time, $\mathcal{D}$ separately decodes the global representation $\tilde{\mathbf{H}}_0$ and the fused sub-category representations $\{\tilde{\mathbf{H}}_i\}_{i=1}^{N}$, producing a primary mask set $\mathcal{M}_0$ together with $N$ sub-category mask sets $\{\mathcal{M}_i\}_{i=1}^{N}$. We then take the union $\bigcup_i \mathcal{M}_i$ and align it with $\mathcal{M}_0$ via mask-level Hungarian matching, attaching a fine-grained sub-category label to each primary mask. For video inference, we adopt the detect-and-track pipeline of SAM~3~\citep{carion2025sam}, where the LVLM is invoked only once per clip and its generated concepts are broadcast to all frames as semantic anchors for per-frame detection, while memory-based mask propagation is delegated to the off-the-shelf tracker without any video-specific modification.

\subsection{Hierarchical Semantic Annotation}
\label{sec:annotation}
Existing referring segmentation datasets mainly focus mainly on instance-level annotation and provide little set-level annotation. To fill this gap, we augment existing reasoning segmentation datasets~\citep{lai2024lisa,ren2024pixellm} with hierarchical semantic annotations through a two-stage pipeline. We adopt Qwen3-VL-235B-A22B~\citep{bai2025qwen3} as the annotator throughout the pipeline for its strong visual grounding and instruction-following ability. In the first stage, starting from the source annotations in which each target is tagged with a category from a closed list, we overlay the instance masks of that category on the image and prompt an LVLM, conditioned on the original label, to produce a free-form sub-category phrase that captures the targets' appearance, attributes, and role in context; the prompt is anchored to the original label to prevent semantic drift. In the second stage, conditioned on the image, the original query, and the collected sub-category phrases, the LVLM is prompted to produce a global set-level concept that is faithful to the query semantics and inclusive of all referred sub-targets. The resulting annotations are further cleaned by rule-based filtering and multiple rounds of manual spot-checking to remove malformed or low-quality cases.

The final corpus contains $236{,}396$ samples and $784{,}809$ concept phrases, averaging $2.32$ sub-categories per sample and reaching up to $16$ in the long tail cases. Over $80\%$ of the samples involve more than one sub-category, supplying the set-level supervision that prior corpora lack. The concept phrases average $3.5$ words, indicating natural-language expressions rather than closed-set labels. The full annotation procedure, filtering criteria, and detailed statistics are provided in Appendix~\ref{sec:appendix_data}.

\section{Experiments}
\subsection{Experimental Setup}
\noindent \textbf{Implementation Details.}
We adopt Qwen3-VL-8B-Instruct~\citep{bai2025qwen3} as the LVLM backbone and SAM~3~\citep{carion2025sam} as the segmentation model. During training, we jointly optimize the multimodal decoder, the projection module, the token embeddings and language-modeling head of the LVLM, and LoRA adapters with rank $128$, $\alpha=256$, and dropout $0.05$ attached to all linear layers of the LVLM language model.  We train the model for one epoch on 8 NVIDIA H200 GPUs using AdamW with a learning rate of $4\times10^{-5}$, a cosine annealing schedule, and a batch size of 64.

\noindent \textbf{Datasets and Benchmarks.}
Our training set comprises both image and video segmentation datasets. For images, we use RefCOCO/+/g~\citep{kazemzadeh2014referitgame, mao2016generation}, gRefCOCO~\citep{liu2023gres}, ReasonSeg~\citep{lai2024lisa}, and MUSE~\citep{ren2024pixellm}; for videos, we use MeViS~\citep{ding2023mevis,ding2025mevis}, Ref-DAVIS~\citep{khoreva2018video}, Ref-YouTube-VOS~\citep{seo2020urvos}, and ReVOS~\citep{yan2024visa}. Referring datasets are used with their original natural-language queries, while reasoning segmentation datasets (ReasonSeg and MUSE) are augmented with hierarchical semantic annotations via the two-stage pipeline in Sec.~\ref{sec:annotation}. We evaluate on the test splits of these training datasets, and additionally report results on ReasonVOS~\citep{bai2024one} and SeCVOS~\citep{zhang2025sec} to assess performance in complex video scenarios. Detailed implementation and evaluation metrics are provided in Appendix~\ref{sec:appendix_setup}.

\subsection{Main Results}

\begin{table*}[t]
    \centering
    \caption{\textbf{Comparison with prior work on multi-object referring segmentation benchmarks}, gRefCOCO~\citep{liu2023gres} and MUSE~\citep{ren2024pixellm}, reporting gIoU, cIoU, and F1@0.5 where available. \modelname obtains the \textbf{best} score on every column. $\dagger$ denotes results reproduced by us.}
    \label{tab:multi_object_seg}
    \resizebox{\textwidth}{!}{
\begin{tabular}{l | cccccc |cccccc}
\toprule
\multirow{4}{*}{\textbf{Model}}
& \multicolumn{6}{c|}{\textbf{gRefCOCO}}
& \multicolumn{6}{c}{\textbf{MUSE}} \\
\cmidrule(lr){2-7}\cmidrule(lr){8-13}
& \multicolumn{2}{c}{val} & \multicolumn{2}{c}{testA} & \multicolumn{2}{c|}{testB}
& \multicolumn{3}{c}{val} & \multicolumn{3}{c}{test} \\
\cmidrule(lr){2-3}\cmidrule(lr){4-5}\cmidrule(lr){6-7}\cmidrule(lr){8-10}\cmidrule(lr){11-13}
& gIoU & cIoU & gIoU & cIoU & gIoU & cIoU & gIoU & cIoU & F1@0.5 & gIoU & cIoU & F1@0.5 \\
\midrule
LISA-7B~\citep{lai2024lisa}                  & \score{61.6} & \score{61.8} & \score{66.3} & \score{68.5} & \score{58.8} & \score{60.6} & \score{42.0} & \score{46.1} & \score{-} & \score{38.9} & \score{44.4} & \score{-} \\
LISA-Llama2-13B~\citep{lai2024lisa}          & \score{63.5} & \score{63.0} & \score{68.2} & \score{69.7} & \score{61.8} & \score{62.2} & \score{43.6} & \score{50.2} & \score{-} & \score{41.9} & \score{50.5} & \score{-}  \\
GSVA~\citep{xia2024gsva}                     & \score{66.5} & \score{63.3} & \score{71.1} & \score{69.9} & \score{62.2} & \score{60.5} & \score{-}    & \score{-} & \score{-} & \score{-}    & \score{-} & \score{-}  \\
PixelLM-7B~\citep{ren2024pixellm}             & \score{-}    & \score{-} & \score{-}    & \score{-} & \score{-}    & \score{-} & \score{42.6} & \score{50.7} & \score{-} & \score{39.2} & \score{46.3} & \score{-}  \\
PixelLM-Llama2-13B~\citep{ren2024pixellm}     & \score{-}    & \score{-} & \score{-}    & \score{-} & \score{-}    & \score{-} & \score{44.8} & \score{54.1} & \score{-} & \score{42.3} & \score{51.0} & \score{-} \\
SAM4MLLM~\citep{chen2024sam4mllm}             & \score{71.9} & \score{67.8} & \score{74.2} & \score{72.2} & \score{65.3} & \score{63.4} & \score{-}    & \score{-} & \score{-} & \score{-}    & \score{-} & \score{-}  \\
Text4Seg~\citep{lan2025text4seg}              & \score{74.4} & \score{69.1} & \score{75.1} & \score{73.8} & \score{67.3} & \score{66.6} & \score{-}    & \score{-} & \score{-} & \score{-}    & \score{-} & \score{-}  \\
MLLMSeg~\citep{wang2025unlocking}             & \score{75.1} & \score{71.6} & \score{77.0} & \score{76.9} & \score{69.7} & \score{68.5} & \score{-}    & \score{-} & \score{-} & \score{-}    & \score{-} & \score{-}  \\
SAM3 Agent~\citep{carion2025sam}              & \scoredag{59.2} & \scoredag{49.1} & \scoredag{63.4} & \scoredag{61.8} & \scoredag{58.5} & \scoredag{55.3} & \scoredag{27.8} & \scoredag{24.2} & \scoredag{41.0} & \scoredag{23.8} & \scoredag{26.3} & \scoredag{38.1}\\
Visionreasoner~\citep{liu2026visionreasoner}  & \score{41.5} & \score{48.3} & \scoredag{58.2} & \scoredag{64.2} & \scoredag{48.5} & \scoredag{51.1} & \scoredag{42.4} & \scoredag{41.6} & \scoredag{54.5} & \scoredag{36.1} & \scoredag{36.8} & \scoredag{49.2} \\
\midrule
\rowcolor{gray!15}
\textbf{\modelname (ours)} & \score{\textbf{78.4}} & \score{\textbf{72.0}} & \score{\textbf{78.5}} & \score{\textbf{78.1}} & \score{\textbf{73.1}} & \score{\textbf{72.4}} & \score{\textbf{56.9}} & \score{\textbf{59.5}} & \score{\textbf{71.2}} & \score{\textbf{53.2}} & \score{\textbf{60.0}} & \score{\textbf{69.3}} \\
\bottomrule
\end{tabular}
    }
    \vspace{-12pt}
\end{table*}

\begin{table*}[t]
    \centering
    \caption{\textbf{Comparison with prior work on referring video object segmentation benchmarks}~\citep{yan2024visa,bai2024one,seo2020urvos,khoreva2018video,ding2023mevis,ding2025mevis,zhang2025sec}, reporting \mjf. \modelname obtains the \textbf{best} score on all seven benchmarks, with substantial gains on the more challenging MeViS and Ref-SeCVOS settings.}
    \label{tab:referring_vos}
    \resizebox{\textwidth}{!}{
        \begin{tabular}{l | ccccccc}
\toprule
\multirow{2}{*}{\textbf{Model}}    & \multicolumn{7}{c}{\mjf}                                                                                  \\
\cmidrule(lr){2-8}
                                      & \textbf{ReVOS} & \textbf{ReasonVOS} & \textbf{Ref-YTVOS} & \textbf{Ref-DAVIS} & \textbf{MeViS v1} & \textbf{MeViS v2} & \textbf{Ref-SeCVOS}  \\
\midrule
LBDT~\citep{ding2022language}              & -                & -                & {49.4}      & {54.1}       & {29.3}     & {25.1}     & -                \\
ReferFormer~\citep{wu2022language}         & {28.1}     & {32.9}     & {62.9}      & {61.1}       & {31.0}     & {26.7}     & -                \\
VLT+TC~\citep{ding2022vlt}                 & -                & -                & {62.7}      & {60.3}       & {35.6}     & {30.1}     & -                \\
HTML~\citep{han2023html}                   & -                & -                & {63.4}      & {62.1}       & -                & -                & -                \\
OnlineRefer~\citep{wu2023onlinerefer}      & -                & {38.7}     & {63.5}      & {64.8}       & {32.3}     & -                & -                \\
LMPM~\citep{ding2023mevis}                 & {26.4}     & -                & -                 & -                  & {37.2}     & {38.3}     & -                \\
SOC~\citep{luo2023soc}                     & -                & {35.9}     & {66.0}      & {64.2}       & -                & -                & -                \\
SgMg~\citep{miao2023spectrum}              & -                & {36.2}     & {65.7}      & {63.3}       & -                & -                & -                \\
TrackGPT~\citep{zhu2023tracking}           & {45.0}     & -                & {59.5}      & {66.5}       & {41.2}     & -                & -                \\
LISA~\citep{lai2024lisa}                   & {40.9}     & {31.1}     & {53.9}      & {64.8}       & {37.2}     & -                & -                \\
VideoLISA~\citep{bai2024one}               & -                & {47.5}     & {63.7}      & {68.8}       & {44.4}     & -                & {42.8}     \\
VISA~\citep{yan2024visa}                   & {50.9}     & -                & {63.0}      & {70.4}       & {44.5}     & -                & {59.5}     \\
DsHmp~\citep{he2024decoupling}             & -                & -                & {67.1}      & {64.9}       & {46.4}     & {40.8}     & -                \\
DMVS~\citep{wang2025deforming}             & -                & -                & {64.3}      & {65.2}       & {48.6}     & -                & -                \\
VideoGLaMM~\citep{munasinghe2025videoglamm}& -                & -                & -                 & {69.5}       & {45.2}     & -                & -                \\
ViLLa~\citep{zheng2025villa}               & {57.0}     & -                & {67.5}      & {74.3}       & {49.4}     & -                & -                \\
SAMWISE~\citep{cuttano2025samwise}         & -                & -                & {69.2}      & {70.6}       & {49.5}     & -                & {54.0}     \\
GLUS~\citep{lin2025glus}                   & {54.9}     & {49.9}     & {67.3}      & {73.9}    & {51.3}     & {46.5}  & {59.8}     \\
Sa2VA~\citep{yuan2025sa2va}                & {60.4}  & {56.1}  & {72.3}      & {75.9}       & {51.5}  & {47.4}  & {51.8}     \\
InstructSeg~\citep{wei2025instructseg}     & {54.5}     & -                & {67.5}      & {71.1}       & -                & -                & -                \\
VRS-HQ~\citep{gong2025devil}               & {59.1}     & -                & {70.4}      & {76.0}       & {50.6}     & -                & -                \\
SDAM~\citep{zhu2026training}               & {58.0}     & {55.1}     & {65.3}      & {76.0}       & {48.6}     & -                & -                \\
\midrule
\rowcolor{gray!15}
\textbf{\modelname (ours)} & {\textbf{69.5}} & {\textbf{68.0}} & {\textbf{78.8}} & {\textbf{80.8}} & {\textbf{62.4}} & {\textbf{60.1}} & {\textbf{72.2}} \\
\bottomrule
\end{tabular}

    }
    \vspace{-6pt}
\end{table*}

\begin{table*}[t]
    \centering
    \begin{minipage}[t]{0.48\textwidth}
        \centering
        \caption{\textbf{Ablation on the proposed modules.} Adding set-level prediction and explicit conditioning each contributes a further improvement.}
        \label{tab:ablation_method}
        \vspace{-6pt}
        \resizebox{\textwidth}{!}{
            \begin{tabular}{x{45} x{55} | x{35} x{50}}
\toprule
\makecell{\textbf{Set-level}\\\textbf{Prediction}} & \makecell{\textbf{Concept}\\\textbf{Condition}} & \textbf{gIoU} & \textbf{F1@0.5} \\
\midrule
\ding{55} & \ding{55} & 43.6 & 59.2 \\
\ding{51} & \ding{55} & 51.4 & 67.8 \\
\ding{51} & \ding{51} & \textbf{53.2} & \textbf{69.3} \\
\bottomrule
\end{tabular}
        }
    \end{minipage}
    \vspace{-6pt}
    \hfill
    \begin{minipage}[t]{0.48\textwidth}
        \centering
        \caption{\textbf{Ablation on the proposed annotations.} Combining the ``Diverse Labeling'' and ``Hierarchical Semantic'' yields the best result.}
        \vspace{-6pt}
        \label{tab:ablation_annotation}
        \resizebox{\textwidth}{!}{
            \begin{tabular}{x{45} x{55} | x{35} x{50}}
\toprule
\makecell{\textbf{Diverse}\\\textbf{Labeling}} & \makecell{\textbf{Hierarchical}\\\textbf{Semantic}} & \textbf{gIoU} & \textbf{LLM Score} \\
\midrule
\ding{55} & \ding{55} & 50.3 & 5.99 \\
\ding{51} & \ding{55} & 50.0 & 6.92 \\
\ding{51} & \ding{51} & \textbf{53.2} & \textbf{6.94} \\
\bottomrule
\end{tabular}

        }
    \end{minipage}
    \vspace{-6pt}
\end{table*}

\begin{table*}[t]
    \centering
    \caption{\textbf{Comparison with prior work on single-object referring and reasoning segmentation benchmarks} RefCOCO/+/g~\citep{kazemzadeh2014referitgame, mao2016generation} and ReasonSeg~\citep{lai2024lisa}. \modelname remains competitive on RefCOCO/+/g, achieves the best results on ReasonSeg, and obtains the best overall average, indicating that the set-level formulation \textbf{does not regress} on conventional single-target settings.}
    \label{tab:single_object_seg}
    \resizebox{\textwidth}{!}{
        \begin{tabular}{y{120} | x{20} x{20} x{20} | x{20} x{20} x{20} | x{20} x{20} | x{20} x{20} | x{20}}
\toprule
\multirow{2}{*}{\textbf{Model}} & \multicolumn{3}{c|}{\textbf{RefCOCO}} & \multicolumn{3}{c|}{\textbf{RefCOCO$+$}} & \multicolumn{2}{c|}{\textbf{RefCOCOg}} & \multicolumn{2}{c|}{\textbf{ReasonSeg}} & \multirow{2}{*}{\textbf{Avg.}} \\
\cmidrule(lr){2-4}\cmidrule(lr){5-7}\cmidrule(lr){8-9}\cmidrule(lr){10-11}
& val & testA & testB & val & testA & testB & val & test & val & test \\
\midrule
LISA~\citep{lai2024lisa}                     & 74.9 & 79.1 & 72.3 & 65.1 & 70.8 & 58.1 & 67.9 & 70.6 & 46.0 & 34.1 & 63.9 \\
PixelLM~\citep{ren2024pixellm}               & 73.0 & 76.5 & 68.2 & 66.3 & 71.7 & 58.3 & 69.3 & 70.5 & -    & -    & -    \\
GSVA~\citep{xia2024gsva}                     & 79.2 & 81.7 & 77.1 & 70.3 & 73.8 & 63.6 & 75.7 & 77.0 & -    & -    & -    \\
GLaMM~\citep{rasheed2024glamm}               & 79.5 & 83.2 & 76.9 & 72.6 & 78.7 & 64.6 & 74.2 & 74.9 & 47.2 & -    & -    \\
SAM4MLLM~\citep{chen2024sam4mllm}            & 79.8 & 82.7 & 74.7 & 74.6 & 80.0 & 67.2 & 75.5 & 76.4 & 60.4 & -    & -    \\
GLEE~\citep{wu2024general}                   & 80.0 & -    & -    & 69.6 & -    & -    & 72.9 & -    & -    & -    & -    \\
UniLSeg~\citep{liu2024universal}             & 81.7 & 83.2 & 79.9 & 73.2 & 78.3 & 68.2 & 79.3 & 80.5 & -    & -    & -    \\
EVF\text{-}SAM~\citep{zhang2024evf}          & 82.4 & 84.2 & 80.2 & 76.5 & 80.0 & 71.9 & 78.2 & 78.3 & -    & -    & -    \\
PSALM~\citep{zhang2024psalm}                 & 83.6 & 84.7 & 81.6 & 72.9 & 75.5 & 70.1 & 73.8 & 74.4 & -    & -    & -    \\
HyperSeg~\citep{wei2024hyperseg}             & 84.8 & 85.7 & 83.4 & 79.0 & 83.5 & 75.2 & 79.4 & 78.9 & 56.7 & -    & -    \\
Text4Seg~\citep{lan2025text4seg}             & 79.2 & 81.7 & 75.6 & 72.8 & 77.9 & 66.5 & 74.0 & 75.3 & -    & -    & -    \\
DETRIS~\citep{huang2025densely}              & 81.0 & 81.9 & 79.0 & 75.2 & 78.6 & 70.2 & 74.6 & 75.3 & -    & -    & -    \\
MLLMSeg~\citep{wang2025unlocking}            & 81.0 & 82.4 & 78.7 & 76.4 & 79.1 & 72.5 & 79.9 & 80.8 & -    & -    & -    \\
Sa2VA~\citep{yuan2025sa2va}                  & 81.6 & -    & -    & 76.2 & -    & -    & 78.7 & -    & -    & -    & -    \\
RICE~\citep{xie2025region}                   & 83.5 & 85.3 & 81.7 & 79.4 & 82.8 & 75.4 & 79.8 & 80.4 & -    & -    & -    \\
X\text{-}SAM~\citep{wang2025xsam}            & 85.1 & 87.1 & 83.4 & 78.0 & 81.0 & 74.4 & 83.8 & 83.9 & 32.9 & 41.0 & 73.1 \\
RSVP~\citep{lu2025rsvp}                      & -    & -    & -    & -    & -    & -    & 65.5 & 66.4 & 56.7 & 50.7 & -    \\
Seg-Zero~\citep{liu2025segzero}              & -    & 80.3 & -    & -    & 76.2 & -    & -    & 72.6 & 62.0 & 52.0 & -    \\
Visionreasoner~\citep{liu2026visionreasoner} & 76.8 & 78.9 & 72.5 & 70.9 & 74.9 & 64.6 & 72.9 & 71.3 & 66.3 & 63.6 & 71.3 \\
SAM3 Agent~\citep{carion2025sam}             & 68.6 & 72.3 & 63.9 & 59.3 & 64.5 & 55.6 & 66.4 & 66.9 & 57.4 & 67.3 & 64.2 \\
\midrule
\rowcolor{gray!15}
\textbf{\modelname (ours)} & {83.7} & {84.4} & {81.6} & {79.4} & {82.3} & {75.6} & {80.9} & {80.0} & {70.6} & {70.5} & \textbf{78.9} \\
\bottomrule
\end{tabular}
    }
    \vspace{-9pt}
\end{table*}

We conduct a comprehensive evaluation of \modelname on $6$ image and $7$ video segmentation benchmarks. The comparison includes both classical segmentation models and recent LVLM-based approaches.

As shown in Tab.~\ref{tab:multi_object_seg}, \modelname achieves superior performance on the multi-object segmentation benchmarks gRefCOCO and MUSE, surpassing the previous state of the art by +12.1 gIoU on MUSE \textit{val} and +3.3 gIoU on gRefCOCO \textit{val}. Notably, as illustrated in Fig.~\ref{fig:preliminary}(a), the performance margin becomes more pronounced as the number of target objects increases. This trend suggests that our explicit semantic set prediction design provides strong stability and scalability in challenging multi-target scenarios. Moreover, emphasizing set-level concepts does not compromise single-target segmentation accuracy. As reported in Tab.~\ref{tab:single_object_seg}, \modelname also achieves the competitive performance across RefCOCO/+/g and ReasonSeg, demonstrating that the proposed set-level formulation remains compatible with conventional single-object referring segmentation.

Tab.~\ref{tab:referring_vos} further demonstrates that \modelname establishes new state-of-the-art results across all seven referring video object segmentation benchmarks. In particular, the improvements are especially pronounced on the more challenging MeViS and Ref-SeCVOS benchmarks, where \modelname surpasses previous best methods by $+10.9$ and $+12.4$ \mjf, respectively. These substantial gains suggest that explicit semantic concepts serve as stable target anchors for long-horizon temporal association and semantically complex video understanding. As a result, \modelname can maintain more robust object tracking under challenging conditions such as occlusion, distractors, and appearance changes. 

\subsection{Ablation Study and Analysis}

\begin{figure}[t]
    \centering
    \includegraphics[width=\textwidth]{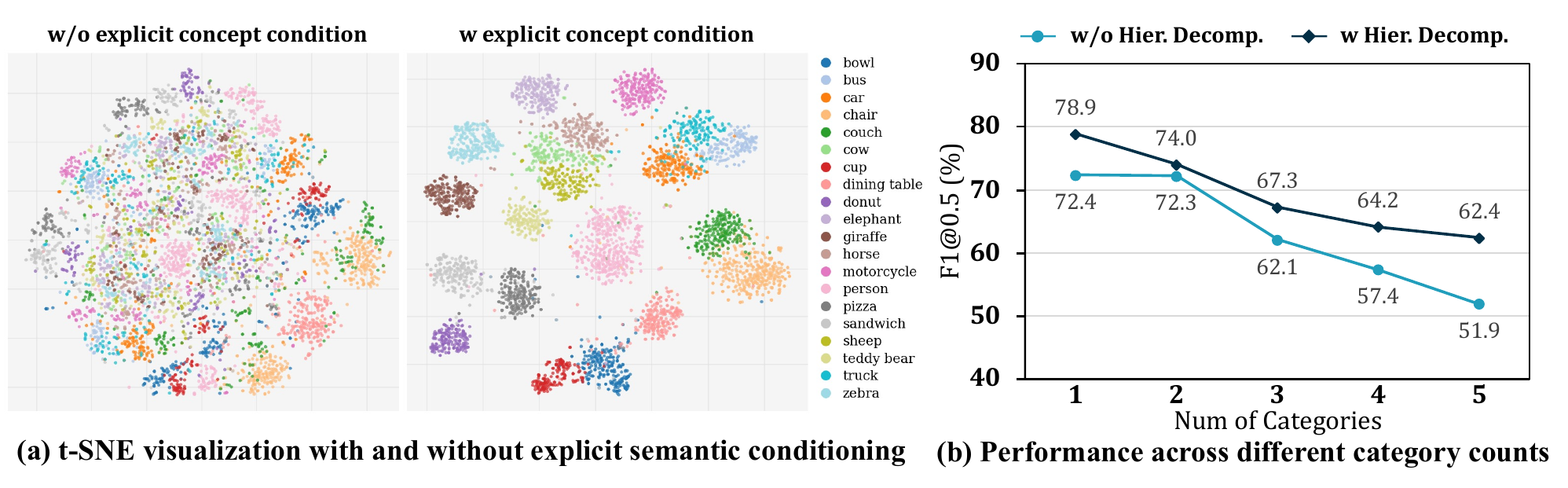}
    \caption{\textbf{Ablation analysis of \modelname on MUSE.} \textbf{(a)} t-SNE projection of segmentation conditions, colored by category: explicit concept conditioning yields tighter, more category-aligned clusters than the token-only variant. \textbf{(b)} F1@0.5 versus the number of referred categories (1--5): hierarchical decomposition consistently outperforms the flat variant, with the gap widening as cardinality grows.}
    \label{fig:ablation_analysis}
    \vspace{-12pt}
\end{figure}

We conduct a series of ablation studies on the MUSE test set to isolate the contributions of our architectural design and annotation quality, with results reported in Tab.~\ref{tab:ablation_method} and Tab.~\ref{tab:ablation_annotation}.

\noindent \textbf{Effectiveness of proposed modules.}
Tab.~\ref{tab:ablation_method} presents an ablation study evaluating the contributions of set-level prediction and explicit concept conditioning. Starting from a baseline that follows the prevailing implicit-token paradigm, introducing set-level prediction yields a substantial improvement, suggesting that modeling referred targets as a set better aligns with the requirements of open-ended scenarios. Incorporating explicit concept conditioning further improves performance, indicating that grounding targets with interpretable semantic concepts provides additional discriminative cues beyond latent mask tokens. We further compare the learned feature spaces of token-based and concept-based representations in Fig.~\ref{fig:ablation_analysis}(a). Compared with special token representations, concept-based representations form clearer concept-level clusters, corroborating that explicit semantic anchoring reshapes the feature space toward semantic structure and makes it better suited to open-ended queries.

\noindent \textbf{Effectiveness of proposed datasets.}
To evaluate the impact of our annotation pipeline, we ablate its two stages and use an LVLM-based judge to assess the quality of the resulting semantic labels. As shown in Tab.~\ref{tab:ablation_annotation}, replacing the original rigid labels with diverse natural-language descriptions improves the semantic expressiveness while preserving mask prediction quality. This indicates that richer descriptions provide more informative semantic cues without compromising localization accuracy. Further adding hierarchical semantic annotation leads to more accurate predictions, suggesting that organizing the enriched labels into a coarse-to-fine structure helps the model better exploit semantic supervision. Fig.~\ref{fig:ablation_analysis}(b) further shows that this advantage becomes more pronounced as the number of referred categories increases, highlighting the benefit of hierarchy-aware supervision for high-cardinality and compositionally complex queries.

\begin{figure}[t]
    \centering
    \includegraphics[width=\textwidth]{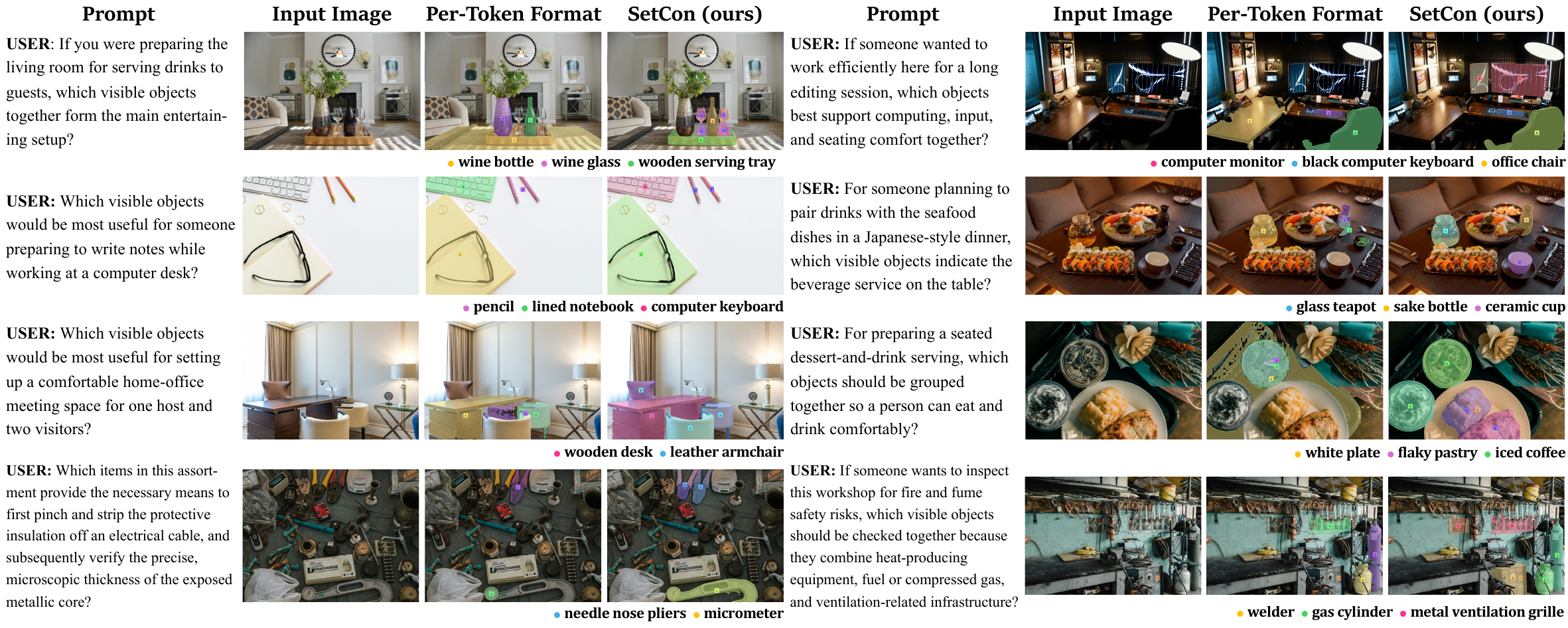}
    \caption{\textbf{Qualitative comparison on in-the-wild images.} We show the user prompt, input image, the LISA-format baseline, and \modelname, with the predicted sub-category concepts listed below. Across reasoning-style queries spanning \textit{multi-instance}, \textit{cross-category}, and \textit{open-ended scenarios}, \modelname tends to produce more complete and semantically coherent mask sets.}
    \label{fig:qualitative}
    \vspace{-6pt}
\end{figure}

\noindent \textbf{Qualitative Results.} To more intuitively demonstrate the segmentation performance of our framework in real-world scenarios, Fig.~\ref{fig:qualitative} presents a visual comparison between \modelname and a standard LISA-format baseline on in-the-wild images. As shown, our method consistently produces more accurate and semantically coherent masks across diverse visual scenes. Beyond simply localizing the referred targets, \modelname can distinguish multiple object instances based on fine-grained category semantics, as well as appearance and contextual attributes. Benefiting from our hierarchical semantic annotations, the model is better able to activate and ground the world knowledge encoded in the LVLM, enabling robust generalization to long-tail, cross-category, and open-ended scenarios. Additional qualitative results on video referring segmentation are provided in Appendix~\ref{sec:appendix_video}.

\begin{figure}[t]
    \centering
    \includegraphics[width=\textwidth]{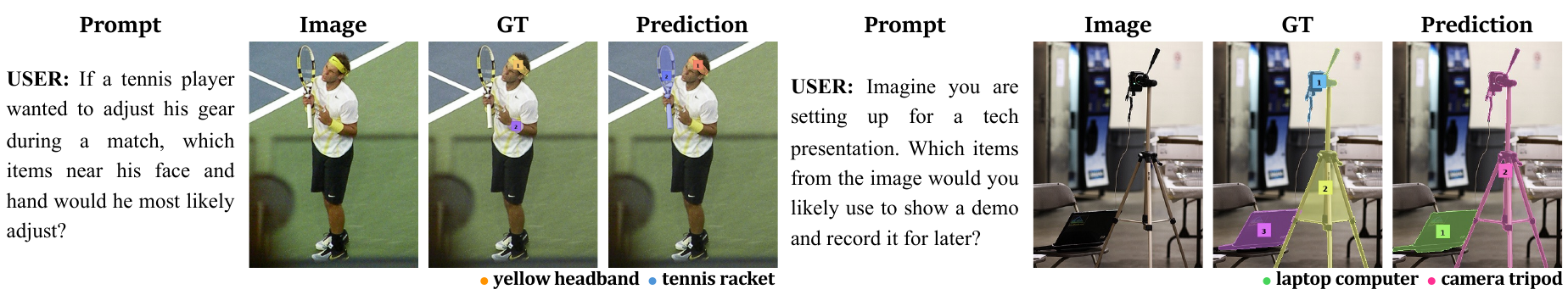}
    \caption{\textbf{Failure cases on the MUSE benchmark.} Open-ended queries expose limitations in target disambiguation and concept granularity, which may lead to incorrect or incomplete predictions.}
    \label{fig:failure_case}
    \vspace{-6pt}
\end{figure}

\noindent \textbf{Failure Cases.} However, since \modelname captures the overall scene semantics, open-ended queries may introduce ambiguity in the intended targets, and the semantic boundaries between related objects can be blurred. As illustrated in Fig.~\ref{fig:failure_case}, open-ended queries may cause ambiguity in target selection and concept granularity. Relational descriptions may refer to multiple plausible objects satisfying the same spatial relation, while functional descriptions may compose related objects into a common semantic concept, leading to mismatches with the annotated target set.

\section{Conclusion}
In this paper, we presented \textbf{Set}-\textbf{Con}cept Segmentation (\textbf{\modelname}), a framework that reformulates open-ended referring segmentation as explicit set-level concept prediction. Through an interpretable language-based concept interface and hierarchical semantic decomposition, \modelname jointly grounds target sets and decodes coherent mask set, enabling effective segmentation in multi-instance, cross-category, and open-ended scenarios. We further construct large-scale hierarchical supervision to enhance concept-to-mask learning and open-world generalization. Extensive experiments on image and video benchmarks show that \modelname achieves leading performance, particularly in challenging multi-target and cross-category settings. We hope this work provides a simple yet effective step toward more semantic, scalable, and general-purpose segmentation systems.

\newpage
\section*{Acknowledgments}
This paper is funded in part by the National Science and Technology Major Project of China (Grant No. 2026ZD0128702) and by the Shanghai Innovation Institute.

{\small
\bibliographystyle{plain}
\bibliography{reference}

@String(CVPR  = {IEEE Conf. Comput. Vis. Pattern Recog.})

@String(ICCV  = {Int. Conf. Comput. Vis.})

@String(ECCV  = {Eur. Conf. Comput. Vis.})

@String(AAAI  = {AAAI})

@String(CVPR  = {CVPR})

@String(ICCV  = {ICCV})

@String(ECCV  = {ECCV})

@inproceedings{lai2024lisa,
  title={Lisa: Reasoning segmentation via large language model},
  author={Lai, Xin and Tian, Zhuotao and Chen, Yukang and Li, Yanwei and Yuan, Yuhui and Liu, Shu and Jia, Jiaya},
  booktitle={Proceedings of the IEEE/CVF conference on computer vision and pattern recognition},
  pages={9579--9589},
  year={2024}
}

@inproceedings{ren2024pixellm,
  title={Pixellm: Pixel reasoning with large multimodal model},
  author={Ren, Zhongwei and Huang, Zhicheng and Wei, Yunchao and Zhao, Yao and Fu, Dongmei and Feng, Jiashi and Jin, Xiaojie},
  booktitle={Proceedings of the IEEE/CVF Conference on Computer Vision and Pattern Recognition},
  pages={26374--26383},
  year={2024}
}

@inproceedings{jang2025mmr,
  title={MMR: A Large-scale Benchmark Dataset for Multi-target and Multi-granularity Reasoning Segmentation},
  author={Jang, Donggon and Cho, Yucheol and Lee, Suin and Kim, Taehyeon and Kim, Daeshik},
  booktitle={The Thirteenth International Conference on Learning Representations},
  year={2025}
}

@inproceedings{liu2026visionreasoner,
  title={VisionReasoner: Unified Reasoning-Integrated Visual Perception via Reinforcement Learning},
  author={Liu, Yuqi and Qu, Tianyuan and Zhong, Zhisheng and Peng, Bohao and Liu, Shu and Yu, Bei and Jia, Jiaya},
  booktitle={The Fourteenth International Conference on Learning Representations},
  year={2026}
}

@inproceedings{carion2025sam,
  title={Sam 3: Segment anything with concepts},
  author={Carion, Nicolas and Gustafson, Laura and Hu, Yuan-Ting and Debnath, Shoubhik and Hu, Ronghang and Suris, Didac and Ryali, Chaitanya and Alwala, Kalyan Vasudev and Khedr, Haitham and Huang, Andrew and others},
  booktitle={The Fourteenth International Conference on Learning Representations},
  year={2026}
}

@inproceedings{xia2024gsva,
  title={Gsva: Generalized segmentation via multimodal large language models},
  author={Xia, Zhuofan and Han, Dongchen and Han, Yizeng and Pan, Xuran and Song, Shiji and Huang, Gao},
  booktitle={Proceedings of the IEEE/CVF Conference on Computer Vision and Pattern Recognition},
  pages={3858--3869},
  year={2024}
}

@inproceedings{zhang2024psalm,
  title={Psalm: Pixelwise segmentation with large multi-modal model},
  author={Zhang, Zheng and Ma, Yeyao and Zhang, Enming and Bai, Xiang},
  booktitle={European Conference on Computer Vision},
  pages={74--91},
  year={2024},
  organization={Springer}
}

@inproceedings{chen2024sam4mllm,
  title={Sam4mllm: Enhance multi-modal large language model for referring expression segmentation},
  author={Chen, Yi-Chia and Li, Wei-Hua and Sun, Cheng and Wang, Yu-Chiang Frank and Chen, Chu-Song},
  booktitle={European Conference on Computer Vision},
  pages={323--340},
  year={2024},
  organization={Springer}
}

@inproceedings{lan2025text4seg,
  title={Text4Seg: Reimagining Image Segmentation as Text Generation},
  author={Lan, Mengcheng and Chen, Chaofeng and Zhou, Yue and Xu, Jiaxing and Ke, Yiping and Wang, Xinjiang and Feng, Litong and Zhang, Wayne},
  booktitle={The Thirteenth International Conference on Learning Representations},
  year={2025}
}

@article{wang2025unlocking,
  title={Unlocking the Potential of MLLMs in Referring Expression Segmentation via a Light-weight Mask Decoder},
  author={Wang, Jingchao and Wu, Zhijian and Huang, Dingjiang and Zheng, Yefeng and Wang, Hong},
  journal={arXiv preprint arXiv:2508.04107},
  year={2025}
}

@article{yuan2025sa2va,
  title={Sa2va: Marrying sam2 with llava for dense grounded understanding of images and videos},
  author={Yuan, Haobo and Li, Xiangtai and Zhang, Tao and Sun, Yueyi and Huang, Zilong and Xu, Shilin and Ji, Shunping and Tong, Yunhai and Qi, Lu and Feng, Jiashi and others},
  journal={arXiv preprint arXiv:2501.04001},
  year={2025}
}

@inproceedings{rasheed2024glamm,
  title={Glamm: Pixel grounding large multimodal model},
  author={Rasheed, Hanoona and Maaz, Muhammad and Shaji, Sahal and Shaker, Abdelrahman and Khan, Salman and Cholakkal, Hisham and Anwer, Rao M and Xing, Eric and Yang, Ming-Hsuan and Khan, Fahad S},
  booktitle={Proceedings of the IEEE/CVF Conference on Computer Vision and Pattern Recognition},
  pages={13009--13018},
  year={2024}
}

@inproceedings{wu2024general,
  title={General object foundation model for images and videos at scale},
  author={Wu, Junfeng and Jiang, Yi and Liu, Qihao and Yuan, Zehuan and Bai, Xiang and Bai, Song},
  booktitle={Proceedings of the IEEE/CVF Conference on Computer Vision and Pattern Recognition},
  pages={3783--3795},
  year={2024}
}

@inproceedings{huang2025densely,
  title={Densely connected parameter-efficient tuning for referring image segmentation},
  author={Huang, Jiaqi and Xu, Zunnan and Liu, Ting and Liu, Yong and Han, Haonan and Yuan, Kehong and Li, Xiu},
  booktitle={Proceedings of the AAAI Conference on Artificial Intelligence},
  pages={3653--3661},
  year={2025}
}

@inproceedings{liu2024universal,
  title={Universal segmentation at arbitrary granularity with language instruction},
  author={Liu, Yong and Zhang, Cairong and Wang, Yitong and Wang, Jiahao and Yang, Yujiu and Tang, Yansong},
  booktitle={Proceedings of the IEEE/CVF Conference on Computer Vision and Pattern Recognition},
  pages={3459--3469},
  year={2024}
}

@article{zhang2024evf,
  title={Evf-sam: Early vision-language fusion for text-prompted segment anything model},
  author={Zhang, Yuxuan and Cheng, Tianheng and Zhu, Lianghui and Hu, Rui and Liu, Lei and Liu, Heng and Ran, Longjin and Chen, Xiaoxin and Liu, Wenyu and Wang, Xinggang},
  journal={arXiv preprint arXiv:2406.20076},
  year={2024}
}

@inproceedings{xie2025region,
  title={Region-based cluster discrimination for visual representation learning},
  author={Xie, Yin and Yang, Kaicheng and An, Xiang and Wu, Kun and Zhao, Yongle and Deng, Weimo and Ran, Zimin and Wang, Yumeng and Feng, Ziyong and Miles, Roy and others},
  booktitle={Proceedings of the IEEE/CVF International Conference on Computer Vision},
  pages={1793--1803},
  year={2025}
}

@InProceedings{wei2024hyperseg,
    author    = {Wei, Cong and Zhong, Yujie and Tan, Haoxian and Liu, Yong and Hu, Jie and Li, Dengjie and Zhao, Zheng and Yang, Yujiu},
    title     = {HyperSeg: Hybrid Segmentation Assistant with Fine-grained Visual Perceiver},
    booktitle = {Proceedings of the Computer Vision and Pattern Recognition Conference (CVPR)},
    month     = {June},
    year      = {2025},
    pages     = {8931-8941}
}

@inproceedings{wang2025xsam,
  title={X-SAM: From segment anything to any segmentation},
  author={Wang, Hao and Qiao, Limeng and Jie, Zequn and Huang, Zhijian and Feng, Chengjian and Zheng, Qingfang and Ma, Lin and Lan, Xiangyuan and Liang, Xiaodan},
  booktitle={Proceedings of the AAAI Conference on Artificial Intelligence},
  pages={26187--26196},
  year={2026}
}

@inproceedings{liu2023gres,
  title={Gres: Generalized referring expression segmentation},
  author={Liu, Chang and Ding, Henghui and Jiang, Xudong},
  booktitle={Proceedings of the IEEE/CVF conference on computer vision and pattern recognition},
  pages={23592--23601},
  year={2023}
}

@inproceedings{ding2023mevis,
  title={Mevis: A large-scale benchmark for video segmentation with motion expressions},
  author={Ding, Henghui and Liu, Chang and He, Shuting and Jiang, Xudong and Loy, Chen Change},
  booktitle={Proceedings of the IEEE/CVF international conference on computer vision},
  pages={2694--2703},
  year={2023}
}

@article{ding2025mevis,
  title={MeViS: A multi-modal dataset for referring motion expression video segmentation},
  author={Ding, Henghui and Liu, Chang and He, Shuting and Ying, Kaining and Jiang, Xudong and Loy, Chen Change and Jiang, Yu-Gang},
  journal={IEEE Transactions on Pattern Analysis and Machine Intelligence},
  year={2025},
  publisher={IEEE}
}

@inproceedings{kazemzadeh2014referitgame,
  title={Referitgame: Referring to objects in photographs of natural scenes},
  author={Kazemzadeh, Sahar and Ordonez, Vicente and Matten, Mark and Berg, Tamara},
  booktitle={Proceedings of the 2014 conference on empirical methods in natural language processing},
  pages={787--798},
  year={2014}
}

@inproceedings{mao2016generation,
  title={Generation and comprehension of unambiguous object descriptions},
  author={Mao, Junhua and Huang, Jonathan and Toshev, Alexander and Camburu, Oana and Yuille, Alan L and Murphy, Kevin},
  booktitle={Proceedings of the IEEE conference on computer vision and pattern recognition},
  pages={11--20},
  year={2016}
}

@inproceedings{zhang2025sec,
  title={Sec: Advancing complex video object segmentation via progressive concept construction},
  author={Zhang, Zhixiong and Ding, Shuangrui and Dong, Xiaoyi and He, Songxin and Lin, Jianfan and Tang, Junsong and Zang, Yuhang and Cao, Yuhang and Lin, Dahua and Wang, Jiaqi},
  booktitle={The Fourteenth International Conference on Learning Representations},
  year={2026}
}

@inproceedings{khoreva2018video,
  title={Video object segmentation with referring expressions},
  author={Khoreva, Anna and Rohrbach, Anna and Schiele, Brent},
  booktitle={Proceedings of the European Conference on Computer Vision (ECCV) Workshops},
  year={2018}
}

@inproceedings{seo2020urvos,
  title={Urvos: Unified referring video object segmentation network with a large-scale benchmark},
  author={Seo, Seonguk and Lee, Joon-Young and Han, Bohyung},
  booktitle={European conference on computer vision},
  pages={208--223},
  year={2020},
}

@inproceedings{yan2024visa,
  title={Visa: Reasoning video object segmentation via large language models},
  author={Yan, Cilin and Wang, Haochen and Yan, Shilin and Jiang, Xiaolong and Hu, Yao and Kang, Guoliang and Xie, Weidi and Gavves, Efstratios},
  booktitle={European Conference on Computer Vision},
  pages={98--115},
  year={2024},
}

@article{bai2024one,
  title={One token to seg them all: Language instructed reasoning segmentation in videos},
  author={Bai, Zechen and He, Tong and Mei, Haiyang and Wang, Pichao and Gao, Ziteng and Chen, Joya and Liu, Lei and Zhang, Zheng and Shou, Mike Z},
  journal={Advances in Neural Information Processing Systems},
  pages={6833--6859},
  year={2024}
}

@article{liu2025segzero,
  title        = {Seg-Zero: Reasoning-Chain Guided  Segmentation via Cognitive Reinforcement},
  author       = {Liu, Yuqi and Peng, Bohao and Zhong, Zhisheng and Yue, Zihao and Lu, Fanbin and Yu, Bei and Jia, Jiaya},
  journal      = {arXiv preprint arXiv:2503.06520},
  year         = {2025}
}

@inproceedings{lu2025rsvp,
  title={Rsvp: Reasoning segmentation via visual prompting and multi-modal chain-of-thought},
  author={Lu, Yi and Cao, Jiawang and Wu, Yongliang and Li, Bozheng and Tang, Licheng and Ji, Yangguang and Wu, Chong and Wu, Jay and Zhu, Wenbo},
  booktitle={Proceedings of the 63rd Annual Meeting of the Association for Computational Linguistics},
  pages={14699--14716},
  year={2025}
}

@inproceedings{ding2022language,
  title={Language-bridged spatial-temporal interaction for referring video object segmentation},
  author={Ding, Zihan and Hui, Tianrui and Huang, Junshi and Wei, Xiaoming and Han, Jizhong and Liu, Si},
  booktitle={Proceedings of the IEEE/CVF conference on computer vision and pattern recognition},
  pages={4964--4973},
  year={2022}
}

@inproceedings{wu2022language,
  title={Language as queries for referring video object segmentation},
  author={Wu, Jiannan and Jiang, Yi and Sun, Peize and Yuan, Zehuan and Luo, Ping},
  booktitle={Proceedings of the IEEE/CVF Conference on Computer Vision and Pattern Recognition},
  pages={4974--4984},
  year={2022}
}

@article{ding2022vlt,
  title={VLT: Vision-language transformer and query generation for referring segmentation},
  author={Ding, Henghui and Liu, Chang and Wang, Suchen and Jiang, Xudong},
  journal={IEEE Transactions on Pattern Analysis and Machine Intelligence},
  pages={7900--7916},
  year={2022},
}

@inproceedings{han2023html,
  title={Html: Hybrid temporal-scale multimodal learning framework for referring video object segmentation},
  author={Han, Mingfei and Wang, Yali and Li, Zhihui and Yao, Lina and Chang, Xiaojun and Qiao, Yu},
  booktitle={Proceedings of the IEEE/CVF International Conference on Computer Vision},
  pages={13414--13423},
  year={2023}
}

@inproceedings{wu2023onlinerefer,
  title={Onlinerefer: A simple online baseline for referring video object segmentation},
  author={Wu, Dongming and Wang, Tiancai and Zhang, Yuang and Zhang, Xiangyu and Shen, Jianbing},
  booktitle={Proceedings of the IEEE/CVF International Conference on Computer Vision},
  pages={2761--2770},
  year={2023}
}

@article{luo2023soc,
  title={Soc: Semantic-assisted object cluster for referring video object segmentation},
  author={Luo, Zhuoyan and Xiao, Yicheng and Liu, Yong and Li, Shuyan and Wang, Yitong and Tang, Yansong and Li, Xiu and Yang, Yujiu},
  journal={Advances in Neural Information Processing Systems},
  pages={26425--26437},
  year={2023}
}

@inproceedings{miao2023spectrum,
  title={Spectrum-guided multi-granularity referring video object segmentation},
  author={Miao, Bo and Bennamoun, Mohammed and Gao, Yongsheng and Mian, Ajmal},
  booktitle={Proceedings of the IEEE/CVF International Conference on Computer Vision},
  pages={920--930},
  year={2023}
}

@article{zhu2023tracking,
  title={Tracking with human-intent reasoning},
  author={Zhu, Jiawen and Cheng, Zhi-Qi and He, Jun-Yan and Li, Chenyang and Luo, Bin and Lu, Huchuan and Geng, Yifeng and Xie, Xuansong},
  journal={arXiv preprint arXiv:2312.17448},
  year={2023}
}

@inproceedings{he2024decoupling,
  title={Decoupling static and hierarchical motion perception for referring video segmentation},
  author={He, Shuting and Ding, Henghui},
  booktitle={Proceedings of the IEEE/CVF Conference on Computer Vision and Pattern Recognition},
  pages={13332--13341},
  year={2024}
}

@inproceedings{munasinghe2025videoglamm,
  title={Videoglamm: A large multimodal model for pixel-level visual grounding in videos},
  author={Munasinghe, Shehan and Gani, Hanan and Zhu, Wenqi and Cao, Jiale and Xing, Eric and Khan, Fahad Shahbaz and Khan, Salman},
  booktitle={Proceedings of the Computer Vision and Pattern Recognition Conference},
  pages={19036--19046},
  year={2025}
}

@inproceedings{zheng2025villa,
  title={Villa: Video reasoning segmentation with large language model},
  author={Zheng, Rongkun and Qi, Lu and Chen, Xi and Wang, Yi and Wang, Kun and Zhao, Hengshuang},
  booktitle={Proceedings of the IEEE/CVF International Conference on Computer Vision},
  pages={23667--23677},
  year={2025}
}

@inproceedings{cuttano2025samwise,
  title={Samwise: Infusing wisdom in sam2 for text-driven video segmentation},
  author={Cuttano, Claudia and Trivigno, Gabriele and Rosi, Gabriele and Masone, Carlo and Averta, Giuseppe},
  booktitle={Proceedings of the Computer Vision and Pattern Recognition Conference},
  pages={3395--3405},
  year={2025}
}

@article{wang2025deforming,
  title={Deforming Videos to Masks: Flow Matching for Referring Video Segmentation},
  author={Wang, Zanyi and Jiang, Dengyang and Li, Liuzhuozheng and Dang, Sizhe and Li, Chengzu and Yang, Harry and Dai, Guang and Wang, Mengmeng and Wang, Jingdong},
  journal={arXiv preprint arXiv:2510.06139},
  year={2025}
}

@inproceedings{lin2025glus,
  title={Glus: Global-local reasoning unified into a single large language model for video segmentation},
  author={Lin, Lang and Yu, Xueyang and Pang, Ziqi and Wang, Yu-Xiong},
  booktitle={Proceedings of the Computer Vision and Pattern Recognition Conference},
  pages={8658--8667},
  year={2025}
}

@inproceedings{wei2025instructseg,
  title={Instructseg: Unifying instructed visual segmentation with multi-modal large language models},
  author={Wei, Cong and Zhong, Yujie and Tan, Haoxian and Zeng, Yingsen and Liu, Yong and Wang, Hongfa and Yang, Yujiu},
  booktitle={Proceedings of the IEEE/CVF International Conference on Computer Vision},
  pages={20193--20203},
  year={2025}
}

@inproceedings{gong2025devil,
  title={The devil is in temporal token: High quality video reasoning segmentation},
  author={Gong, Sitong and Zhuge, Yunzhi and Zhang, Lu and Yang, Zongxin and Zhang, Pingping and Lu, Huchuan},
  booktitle={Proceedings of the IEEE/CVF Conference on Computer Vision and Pattern Recognition},
  pages={29183--29192},
  year={2025}
}

@inproceedings{zhu2026training,
  title={Training-Free Spatio-temporal Decoupled Reasoning Video Segmentation with Adaptive Object Memory},
  author={Zhu, Zhengtong and Fan, Jiaqing and Liu, Zhixuan and Li, Fanzhang},
  booktitle={Proceedings of the AAAI Conference on Artificial Intelligence},
  pages={14022--14030},
  year={2026}
}

@article{bai2025qwen3,
  title={Qwen3-vl technical report},
  author={Bai, Shuai and Cai, Yuxuan and Chen, Ruizhe and Chen, Keqin and Chen, Xionghui and Cheng, Zesen and Deng, Lianghao and Ding, Wei and Gao, Chang and Ge, Chunjiang and others},
  journal={arXiv preprint arXiv:2511.21631},
  year={2025}
}

@inproceedings{carion2020detr,
  title={End-to-end object detection with transformers},
  author={Carion, Nicolas and Massa, Francisco and Synnaeve, Gabriel and Usunier, Nicolas and Kirillov, Alexander and Zagoruyko, Sergey},
  booktitle={European conference on computer vision},
  pages={213--229},
  year={2020},
}

@inproceedings{li2023gligen,
  title={Gligen: Open-set grounded text-to-image generation},
  author={Li, Yuheng and Liu, Haotian and Wu, Qingyang and Mu, Fangzhou and Yang, Jianwei and Gao, Jianfeng and Li, Chunyuan and Lee, Yong Jae},
  booktitle={Proceedings of the IEEE/CVF conference on computer vision and pattern recognition},
  pages={22511--22521},
  year={2023}
}

@inproceedings{tu2025videoanydoor,
  title={Videoanydoor: High-fidelity video object insertion with precise motion control},
  author={Tu, Yuanpeng and Luo, Hao and Chen, Xi and Ji, Sihui and Bai, Xiang and Zhao, Hengshuang},
  booktitle={Proceedings of the Special Interest Group on Computer Graphics and Interactive Techniques Conference Conference Papers},
  pages={1--11},
  year={2025}
}

@inproceedings{driess2023palm,
  title={PaLM-E: an embodied multimodal language model},
  author={Driess, Danny and Xia, Fei and Sajjadi, Mehdi SM and Lynch, Corey and Chowdhery, Aakanksha and Ichter, Brian and Wahid, Ayzaan and Tompson, Jonathan and Vuong, Quan and Yu, Tianhe and others},
  booktitle={Proceedings of the 40th International Conference on Machine Learning},
  pages={8469--8488},
  year={2023}
}

@inproceedings{zitkovich2023rt,
  title={Rt-2: Vision-language-action models transfer web knowledge to robotic control},
  author={Zitkovich, Brianna and Yu, Tianhe and Xu, Sichun and Xu, Peng and Xiao, Ted and Xia, Fei and Wu, Jialin and Wohlhart, Paul and Welker, Stefan and Wahid, Ayzaan and others},
  booktitle={Conference on Robot Learning},
  pages={2165--2183},
  year={2023},
}

@inproceedings{yu2018mattnet,
  title={Mattnet: Modular attention network for referring expression comprehension},
  author={Yu, Licheng and Lin, Zhe and Shen, Xiaohui and Yang, Jimei and Lu, Xin and Bansal, Mohit and Berg, Tamara L},
  booktitle={Proceedings of the IEEE conference on computer vision and pattern recognition},
  pages={1307--1315},
  year={2018}
}

@inproceedings{kirillov2023segment,
  title={Segment anything},
  author={Kirillov, Alexander and Mintun, Eric and Ravi, Nikhila and Mao, Hanzi and Rolland, Chloe and Gustafson, Laura and Xiao, Tete and Whitehead, Spencer and Berg, Alexander C and Lo, Wan-Yen and others},
  booktitle={Proceedings of the IEEE/CVF international conference on computer vision},
  pages={4015--4026},
  year={2023}
}

@inproceedings{ravi2025sam2,
  title={SAM 2: Segment Anything in Images and Videos},
  author={Ravi, Nikhila and Gabeur, Valentin and Hu, Yuan-Ting and Hu, Ronghang and Ryali, Chaitanya and Ma, Tengyu and Khedr, Haitham and R{\"a}dle, Roman and Rolland, Chloe and Gustafson, Laura and others},
  booktitle={The Thirteenth International Conference on Learning Representations},
  year={2025}
}

@inproceedings{hu2016segmentation,
  title={Segmentation from natural language expressions},
  author={Hu, Ronghang and Rohrbach, Marcus and Darrell, Trevor},
  booktitle={European conference on computer vision},
  pages={108--124},
  year={2016},
}

@inproceedings{liu2017recurrent,
  title={Recurrent multimodal interaction for referring image segmentation},
  author={Liu, Chenxi and Lin, Zhe and Shen, Xiaohui and Yang, Jimei and Lu, Xin and Yuille, Alan},
  booktitle={Proceedings of the IEEE international conference on computer vision},
  pages={1271--1280},
  year={2017}
}

@inproceedings{yang2022lavt,
  title={Lavt: Language-aware vision transformer for referring image segmentation},
  author={Yang, Zhao and Wang, Jiaqi and Tang, Yansong and Chen, Kai and Zhao, Hengshuang and Torr, Philip HS},
  booktitle={Proceedings of the IEEE/CVF conference on computer vision and pattern recognition},
  pages={18155--18165},
  year={2022}
}

@article{zou2023segment,
  title={Segment everything everywhere all at once},
  author={Zou, Xueyan and Yang, Jianwei and Zhang, Hao and Li, Feng and Li, Linjie and Wang, Jianfeng and Wang, Lijuan and Gao, Jianfeng and Lee, Yong Jae},
  journal={Advances in neural information processing systems},
  volume={36},
  pages={19769--19782},
  year={2023}
}

@article{zhang2024omg,
  title={Omg-llava: Bridging image-level, object-level, pixel-level reasoning and understanding},
  author={Zhang, Tao and Li, Xiangtai and Fei, Hao and Yuan, Haobo and Wu, Shengqiong and Ji, Shunping and Loy, Chen Change and Yan, Shuicheng},
  journal={Advances in neural information processing systems},
  volume={37},
  pages={71737--71767},
  year={2024}
}

@article{singh2025openai,
  title={Openai gpt-5 system card},
  author={Singh, Aaditya and Fry, Adam and Perelman, Adam and Tart, Adam and Ganesh, Adi and El-Kishky, Ahmed and McLaughlin, Aidan and Low, Aiden and Ostrow, AJ and Ananthram, Akhila and others},
  journal={arXiv preprint arXiv:2601.03267},
  year={2025}
}

@article{team2024gemini,
  title={Gemini 1.5: Unlocking multimodal understanding across millions of tokens of context},
  author={Team, Gemini and Georgiev, Petko and Lei, Ving Ian and Burnell, Ryan and Bai, Libin and Gulati, Anmol and Tanzer, Garrett and Vincent, Damien and Pan, Zhufeng and Wang, Shibo and others},
  journal={arXiv preprint arXiv:2403.05530},
  year={2024}
}

@inproceedings{gu2024anomalygpt,
  title={Anomalygpt: Detecting industrial anomalies using large vision-language models},
  author={Gu, Zhaopeng and Zhu, Bingke and Zhu, Guibo and Chen, Yingying and Tang, Ming and Wang, Jinqiao},
  booktitle={Proceedings of the AAAI conference on artificial intelligence},
  pages={1932--1940},
  year={2024}
}

@article{xiu2025viddar,
  title={ViDDAR: Vision language model-based task-detrimental content detection for augmented reality},
  author={Xiu, Yanming and Scargill, Tim and Gorlatova, Maria},
  journal={IEEE transactions on visualization and computer graphics},
  year={2025},
  publisher={IEEE}
}

@inproceedings{hu2023beyond,
  title={Beyond one-to-one: Rethinking the referring image segmentation},
  author={Hu, Yutao and Wang, Qixiong and Shao, Wenqi and Xie, Enze and Li, Zhenguo and Han, Jungong and Luo, Ping},
  booktitle={Proceedings of the IEEE/CVF International Conference on Computer Vision},
  pages={4067--4077},
  year={2023}
}

@inproceedings{wang2024unveiling,
  title={Unveiling parts beyond objects: Towards finer-granularity referring expression segmentation},
  author={Wang, Wenxuan and Yue, Tongtian and Zhang, Yisi and Guo, Longteng and He, Xingjian and Wang, Xinlong and Liu, Jing},
  booktitle={Proceedings of the IEEE/CVF Conference on Computer Vision and Pattern Recognition},
  pages={12998--13008},
  year={2024}
}

@inproceedings{luddecke2022image,
  title={Image segmentation using text and image prompts},
  author={L{\"u}ddecke, Timo and Ecker, Alexander},
  booktitle={Proceedings of the IEEE/CVF conference on computer vision and pattern recognition},
  pages={7086--7096},
  year={2022}
}

@inproceedings{li2022grounded,
  title={Grounded language-image pre-training},
  author={Li, Liunian Harold and Zhang, Pengchuan and Zhang, Haotian and Yang, Jianwei and Li, Chunyuan and Zhong, Yiwu and Wang, Lijuan and Yuan, Lu and Zhang, Lei and Hwang, Jenq-Neng and others},
  booktitle={Proceedings of the IEEE/CVF conference on computer vision and pattern recognition},
  pages={10965--10975},
  year={2022}
}

@inproceedings{liu2024grounding,
  title={Grounding dino: Marrying dino with grounded pre-training for open-set object detection},
  author={Liu, Shilong and Zeng, Zhaoyang and Ren, Tianhe and Li, Feng and Zhang, Hao and Yang, Jie and Jiang, Qing and Li, Chunyuan and Yang, Jianwei and Su, Hang and others},
  booktitle={European conference on computer vision},
  pages={38--55},
  year={2024},
}

@inproceedings{zhong2022regionclip,
  title={Regionclip: Region-based language-image pretraining},
  author={Zhong, Yiwu and Yang, Jianwei and Zhang, Pengchuan and Li, Chunyuan and Codella, Noel and Li, Liunian Harold and Zhou, Luowei and Dai, Xiyang and Yuan, Lu and Li, Yin and others},
  booktitle={Proceedings of the IEEE/CVF conference on computer vision and pattern recognition},
  pages={16793--16803},
  year={2022}
}

@inproceedings{liang2023open,
  title={Open-vocabulary semantic segmentation with mask-adapted clip},
  author={Liang, Feng and Wu, Bichen and Dai, Xiaoliang and Li, Kunpeng and Zhao, Yinan and Zhang, Hang and Zhang, Peizhao and Vajda, Peter and Marculescu, Diana},
  booktitle={Proceedings of the IEEE/CVF conference on computer vision and pattern recognition},
  pages={7061--7070},
  year={2023}
}

@inproceedings{zou2023generalized,
  title={Generalized decoding for pixel, image, and language},
  author={Zou, Xueyan and Dou, Zi-Yi and Yang, Jianwei and Gan, Zhe and Li, Linjie and Li, Chunyuan and Dai, Xiyang and Behl, Harkirat and Wang, Jianfeng and Yuan, Lu and others},
  booktitle={Proceedings of the IEEE/CVF conference on computer vision and pattern recognition},
  pages={15116--15127},
  year={2023}
}

@inproceedings{kamath2021mdetr,
  title={Mdetr-modulated detection for end-to-end multi-modal understanding},
  author={Kamath, Aishwarya and Singh, Mannat and LeCun, Yann and Synnaeve, Gabriel and Misra, Ishan and Carion, Nicolas},
  booktitle={Proceedings of the IEEE/CVF international conference on computer vision},
  pages={1780--1790},
  year={2021}
}

@inproceedings{minderer2022simple,
  title={Simple open-vocabulary object detection},
  author={Minderer, Matthias and Gritsenko, Alexey and Stone, Austin and Neumann, Maxim and Weissenborn, Dirk and Dosovitskiy, Alexey and Mahendran, Aravindh and Arnab, Anurag and Dehghani, Mostafa and Shen, Zhuoran and others},
  booktitle={European conference on computer vision},
  pages={728--755},
  year={2022},
}

@inproceedings{ding2025sam2long,
  title={Sam2long: Enhancing sam 2 for long video segmentation with a training-free memory tree},
  author={Ding, Shuangrui and Qian, Rui and Dong, Xiaoyi and Zhang, Pan and Zang, Yuhang and Cao, Yuhang and Guo, Yuwei and Lin, Dahua and Wang, Jiaqi},
  booktitle={Proceedings of the IEEE/CVF International Conference on Computer Vision},
  pages={13614--13624},
  year={2025}
}

@inproceedings{ding2022motion,
  title={Motion-aware contrastive video representation learning via foreground-background merging},
  author={Ding, Shuangrui and Li, Maomao and Yang, Tianyu and Qian, Rui and Xu, Haohang and Chen, Qingyi and Wang, Jue and Xiong, Hongkai},
  booktitle={2022 IEEE/CVF Conference on Computer Vision and Pattern Recognition (CVPR)},
  pages={9706--9716},
  year={2022},
  organization={IEEE}
}

@inproceedings{ding2024betrayed,
  title={Betrayed by attention: A simple yet effective approach for self-supervised video object segmentation},
  author={Ding, Shuangrui and Qian, Rui and Xu, Haohang and Lin, Dahua and Xiong, Hongkai},
  booktitle={European Conference on Computer Vision},
  pages={215--233},
  year={2024},
  organization={Springer}
}

@inproceedings{qian2023semantics,
  title={Semantics meets temporal correspondence: Self-supervised object-centric learning in videos},
  author={Qian, Rui and Ding, Shuangrui and Liu, Xian and Lin, Dahua},
  booktitle={2023 IEEE/CVF International Conference on Computer Vision (ICCV)},
  pages={16629--16641},
  year={2023},
  organization={IEEE}
}

@inproceedings{ding2023prune,
  title={Prune spatio-temporal tokens by semantic-aware temporal accumulation},
  author={Ding, Shuangrui and Zhao, Peisen and Zhang, Xiaopeng and Qian, Rui and Xiong, Hongkai and Tian, Qi},
  booktitle={2023 IEEE/CVF International Conference on Computer Vision (ICCV)},
  pages={16899--16910},
  year={2023},
  organization={IEEE}
}

@article{ding2026beyond,
  title={Beyond the Current Observation: Evaluating Multimodal Large Language Models in Controllable Non-Markov Games},
  author={Ding, Shengyuan and Wei, Xilin and Fang, Xinyu and Duan, Haodong and Lin, Dahua and Wang, Jiaqi and Zang, Yuhang},
  journal={arXiv preprint arXiv:2606.19338},
  year={2026}
}

@inproceedings{ding2026arm,
  title={Arm-thinker: Reinforcing multimodal generative reward models with agentic tool use and visual reasoning},
  author={Ding, Shengyuan and Fang, Xinyu and Liu, Ziyu and Zang, Yuhang and Cao, Yuhang and Zhao, Xiangyu and Duan, Haodong and Dong, Xiaoyi and Liang, Jianze and Wang, Bin and others},
  booktitle={Proceedings of the IEEE/CVF Conference on Computer Vision and Pattern Recognition},
  pages={22195--22205},
  year={2026}
}

@article{han2026lomo,
  title={LoMo: Local Modality Substitution for Deeper Vision-Language Fusion},
  author={Han, Feng and Zhang, Zhixiong and Liang, Zheming and Wang, Yibin and Wang, Jiaqi},
  journal={arXiv preprint arXiv:2605.30265},
  year={2026}
}

@inproceedings{xing2026caprl,
  title={Caprl: Stimulating dense image caption capabilities via reinforcement learning},
  author={Xing, Long and Dong, Xiaoyi and Zang, Yuhang and Cao, Yuhang and Liang, Jianze and Huang, Qidong and Wang, Jiaqi and Wu, Feng and Lin, Dahua},
  booktitle={International Conference on Learning Representations},
  volume={2026},
  pages={13066--13093},
  year={2026}
}

@article{abtahizadeh2026muse,
  title={Muse Glimmer 30B: An On-Device Agentic Model},
  author={Abtahizadeh, Arash Ebrahim and Aghamelu, Muna and Ahmed, Adil and Acuna, Juan C and Audran-Reiss, Alexis and Avrahamy, Asaf and Bachrach, Yoram and Balioglu, Can and Budhiraja, Amar and {\c{C}}elebi, Onur and others},
  journal={arXiv preprint arXiv:2610.24628},
  year={2026}
}
}

\newpage
\appendix

\section{Additional Qualitative Results on Video}
\label{sec:appendix_video}

\begin{figure}[h]
    \centering
    \vspace{-12pt}
    \includegraphics[width=\textwidth]{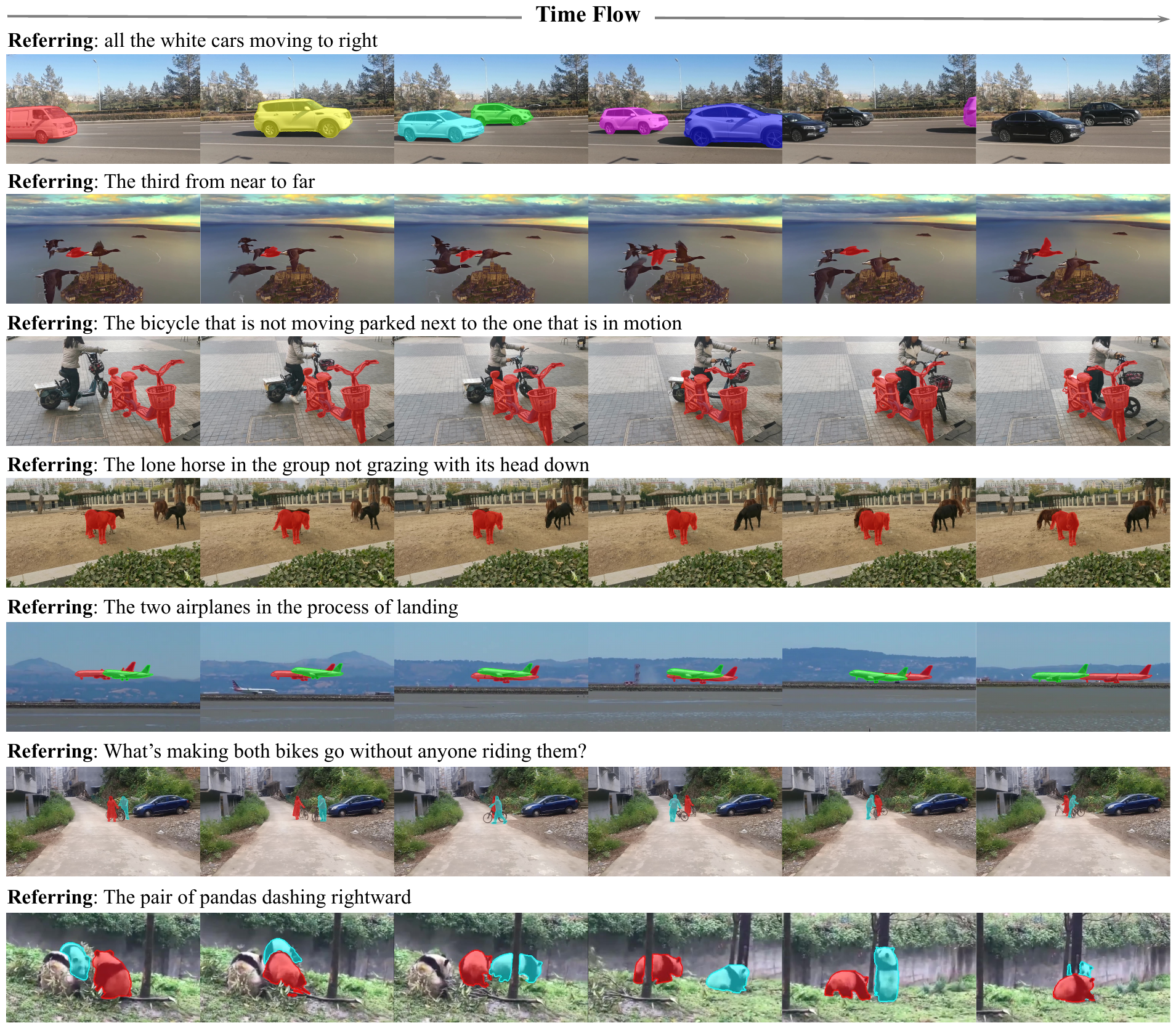}
    \caption{\textbf{Additional qualitative results on referring video object segmentation.} \modelname maintains stable target identities and temporally consistent masks across challenging video sequences with occlusion, distractors, and appearance changes, by using explicit semantic concepts as persistent anchors shared across frames.}
    \vspace{-3pt}
    \label{fig:videovissup}
\end{figure}

We provide additional qualitative results on referring video object segmentation in Fig.~\ref{fig:videovissup}, sampled from the challenging MeViS v2~\citep{ding2025mevis}. The selected clips cover scenarios that are typically challenging for referring video segmentation, including long temporal horizons spanning hundreds of frames, fast or non-rigid object motion, and queries that refer to multiple targets simultaneously. \modelname generalizes naturally to these settings: a single set-level concept is generated once per clip and shared across all frames as a persistent semantic anchor, yielding mask sets that remain temporally coherent and identity-consistent even when individual frames are visually ambiguous in isolation.

\section{Details of Hierarchical Semantic Annotation}
\label{sec:appendix_data}

To support the language-grounded concept interface of \modelname, we augment existing reasoning segmentation datasets~\citep{lai2024lisa,ren2024pixellm} with hierarchical semantic supervision. The pipeline consumes an image, the original natural-language query, the per-target instance masks, and the source closed-vocabulary labels, and produces \textbf{(i)} a free-form sub-category phrase for each target and \textbf{(ii)} a global set-level concept summarizing the target scope. We use \texttt{Qwen3-VL-235B-A22B}~\citep{bai2025qwen3} as the annotator throughout and run it as a two-stage pipeline followed by multi-step quality control.

\noindent \textbf{Stage 1: Diverse Sub-category Labeling.}
For each target, the annotator is shown two views, the original image and the same image overlaid with a colored mask highlighting the target, and is conditioned on the source closed-vocabulary label. It is instructed to produce a natural noun phrase of $1$--$6$ words that may include visual attributes (e.g., color, material) when these are clearly visible, with the goal of replacing rigid category names (e.g., \textit{light}) by more contextually accurate phrases (e.g., \textit{traffic light}, \textit{red bicycle}). The source label is supplied explicitly as an anchor to suppress semantic drift, so that the model refines, but never silently invents, a new category. We sample at a low temperature of $0.2$ to favor faithfulness over diversity.

\noindent \textbf{Stage 2: Set-level Concept Synthesis.}
Given the image, the original query, and the sub-category phrases produced by Stage~1, the annotator is asked to summarize the overall scope of the referred targets into a single \emph{set-level concept}, i.e., an $8$--$15$ word free-form description of the scenario, behavior, or purpose that the targets collectively express (e.g., \textit{the essentials for a carefree day by the water under strong sun}). The output is constrained to lie within $[5, 20]$ words and is truncated when over the upper bound; outputs falling below the lower bound are replaced by a query-conditioned fallback template. We sample at a moderate temperature of $0.7$ to encourage linguistic diversity. The set-level concept produced here, together with the per-target phrases from Stage~1, fully specifies the hierarchical supervision used by \modelname.

\begin{figure}[t]
    \centering
    \includegraphics[width=0.9\textwidth]{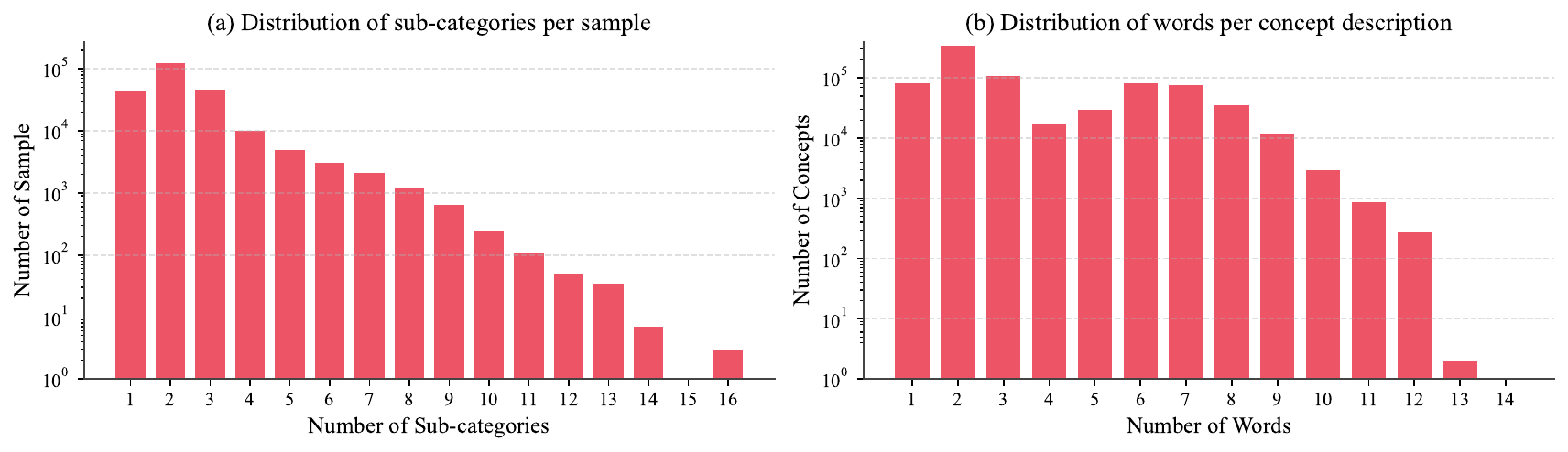}
    \caption{\textbf{Statistics of the training corpus} produced by the proposed two-stage hierarchical annotation pipeline, including the per-sample distribution of sub-categories, and the length distribution of concept phrases.}
    \vspace{-15pt}
    \label{fig:data_stats}
\end{figure}

\noindent \textbf{Quality Control.}
The two annotation stages already enforce light in-line cleaning, including stripping of Markdown markup, URLs, quotes, and boilerplate prefixes (e.g., \textit{The answer is:}), as well as length capping and rejection of overly generic phrases (\eg \textit{object, thing, item}). On top of this, we run a three-step post-processing pipeline that explicitly targets the failure modes we observed in practice:
\begin{itemize}[leftmargin=*,topsep=2pt,itemsep=1pt]
    \item \textbf{Label sanity check.} We identify broken sub-category phrases through two complementary scans. A rule-based pass first catches unambiguous degeneracies, including system-level error tokens (e.g., \textit{OMPI}, \textit{EPHIR}, \textit{MPI\_Init}), residual Markdown markup, URLs, leading exclamation marks, and outputs shorter than two characters. The remaining unique entries are then aggregated into a global vocabulary and reviewed by a text-only LLM judge, which flags non-English characters, gibberish, code snippets, full sentences, apologetic or meta-commentary phrases, and disallowed punctuation. Flagged entries are collected into a global blacklist for the next step.
    \item \textbf{Targeted re-generation.} For each sample whose sub-category phrases include a blacklisted entry, we re-invoke the visual annotator with the same image and mask overlay used in Stage~1 to regenerate, rather than wholesale reverting them to the source label. This step preserves the linguistic richness of Stage~1 wherever the original output was salvageable.
    \item \textbf{Within-sample de-duplication.} A common failure mode is that two distinct sub-category groups under the same query are independently relabeled to the same phrase (e.g., \textit{two ``coffee cup'' groups on the same table}). For each detected duplicate, the annotator is asked to decide, given the image and the two highlighted mask groups, whether to \emph{merge} them into a single sub-category or \emph{split} them into two distinguishable ones. Merging applies when the source dataset has over-split a coherent semantic group, or when the two groups in fact share the same fine-grained semantics; splitting applies when the two groups are visually or functionally distinct but happen to share the same closed-vocabulary label. In the former case, the two mask groups are consolidated under one shared label; in the latter, the duplicate labels are replaced with distinctly worded alternatives. A second de-duplication pass is run afterwards to eliminate any residual collisions.
\end{itemize}
We additionally conduct multiple rounds of manual spot-checking on random subsets to verify annotation faithfulness; systematic failure modes observed during these rounds prompted iterative refinement of the prompts and filters above.

\noindent \textbf{Data Statistics.}
The final corpus contains $236{,}396$ samples and $784{,}809$ concept phrases, with an average of $2.32$ sub-categories per sample and a heavy tail reaching up to $16$ categories in the most complex cases. Over $80\%$ of the samples involve more than one sub-category, supplying the set-level supervision that prior referring corpora lack, and the average concept phrase has $3.5$ words, confirming that the resulting labels behave as natural-language descriptions rather than closed-vocabulary tokens. Detailed distributions over the per-sample number of sub-categories and the length of concept phrases are summarized in Fig.~\ref{fig:data_stats}.

\section{Evaluation Details}
\label{sec:appendix_setup}
For RefCOCO/+/g and ReasonSeg, we use the highest-confidence prediction as the final mask and report cIoU. For the multi-object benchmarks gRefCOCO and MUSE, we discard predicted masks whose confidence is below $0.7$ before aggregation: for gRefCOCO, following~\citep{liu2023gres}, we merge the retained masks into a single foreground mask and report gIoU and cIoU; for MUSE, we align the retained mask set with the ground truth via Hungarian matching, report gIoU and cIoU for segmentation quality, and additionally report F1@0.5 for set-level detection performance. For video benchmarks, we track each predicted target over time, take the union of all tracked masks in each frame as the final foreground sequence, and report standard \mjf.

For the annotation-quality evaluation in Tab.~\ref{tab:ablation_annotation}, we adopt \texttt{Qwen3-VL-8B-Instruct}~\citep{bai2025qwen3} as the judge model. The LLM Score evaluates how accurately and descriptively a predicted label represents its corresponding visual region. For each predicted mask, we match it to the labeled GT mask, highlight the matched region in the original image using a bounding box, and provide the judge with the image, predicted label, and the dataset's original closed-vocabulary label as a semantic reference. The scoring scale and evaluation criteria are defined as follows:

\begin{itemize}[leftmargin=*,topsep=2pt,itemsep=1pt]
    \item \textbf{Scale}: The score ranges from 1 to 10, with 6 representing a correct but generic label.
    \item \textbf{Scoring rubric}: 9--10 --- highly specific, rich, and visually grounded details; 7--8 --- correct label with additional attributes, states, functions, relations, or scene context; 6 --- correct subject with no additional visual detail; 4--5 --- visually related to or confused with a similar category; 2--3 --- incorrect object with some scene or attribute association; 1 --- unrelated to the highlighted region.
\end{itemize}

\section{Additional Experiments}
\label{sec:additional_experiments}

\begin{table}[t]
    \centering
    \begin{minipage}[t]{0.50\textwidth}
        \centering
        \caption{\textbf{Robustness to segmentation backends.} \modelname consistently outperforms previous methods across different segmentation backends.}
        \label{tab:decoder_ablation}
        \vspace{2pt}
        \resizebox{\linewidth}{!}{%
            \begin{tabular}{lccc|ccc}
                \toprule
                \multirow{2}{*}{\textbf{Model}} & \multicolumn{3}{c|}{\textbf{MUSE val}} & \multicolumn{3}{c}{\textbf{MUSE test}} \\
                \cmidrule(lr){2-4}\cmidrule(lr){5-7}
                & \textbf{gIoU} & \textbf{cIoU} & \textbf{F1@0.5} & \textbf{gIoU} & \textbf{cIoU} & \textbf{F1@0.5} \\
                \midrule
                PixelLM & 42.6 & 50.7 & -- & 39.2 & 46.3 & -- \\
                SAM 3 Agent & 27.8 & 24.2 & 41.0 & 23.8 & 26.3 & 38.1 \\
                VisionReasoner & 42.4 & 41.6 & 54.5 & 36.1 & 36.8 & 49.2 \\
                \midrule
                \rowcolor{gray!10}
                \modelname (GroundingDINO) & 50.6 & 51.6 & 63.6 & 50.1 & 56.6 & 64.4 \\
                \rowcolor{gray!15}
                \textbf{\modelname (SAM 3)} & \textbf{56.9} & \textbf{59.5} & \textbf{71.2} & \textbf{53.2} & \textbf{60.0} & \textbf{69.3} \\
                \bottomrule
            \end{tabular}
        }
    \end{minipage}\hfill
    \begin{minipage}[t]{0.48\textwidth}
        \centering
        \caption{\textbf{Robustness to judge models.} Diverse labeling and hierarchical semantics yield consistent gains across all judges.}
        \label{tab:judge_ablation}
        \vspace{2pt}
        \resizebox{\linewidth}{!}{%
            \begingroup
\scriptsize
\setlength{\tabcolsep}{1.5pt}
\renewcommand{\arraystretch}{1.08}
\begin{tabular}{@{}cc|ccccc@{}}
\toprule
\makecell{\textbf{Div.}\\\textbf{Label}} &
\makecell{\textbf{Hier.}\\\textbf{Sem.}} &
\textbf{Qwen3-VL} &
\textbf{GPT-5.5} &
\textbf{Gemini} &
\textbf{Claude} &
\textbf{Doubao} \\
\midrule
\ding{55} & \ding{55} & 5.99 & 5.99 & 5.75 & 5.97 & 5.80 \\
\ding{51} & \ding{55} & 6.89 & 6.71 & 6.57 & 6.80 & 6.40 \\
\ding{51} & \ding{51} & \textbf{6.93} & \textbf{6.76} & \textbf{6.85} & \textbf{6.86} & \textbf{6.69} \\
\bottomrule
\end{tabular}
\endgroup

        }
    \end{minipage}
\end{table}

\subsection{Robustness to Segmentation Backends}
\label{sec:decoder_ablation}
To investigate whether \modelname relies on a specific segmentation backend, we replace SAM~3 with GroundingDINO~\citep{liu2024grounding} while keeping the remaining pipeline unchanged. Since GroundingDINO produces bounding boxes rather than segmentation masks, we use SAM to convert the detected boxes into masks. As shown in Tab.~\ref{tab:decoder_ablation}, \modelname with GroundingDINO still consistently outperforms previous methods. This demonstrates that the improvements mainly originate from the proposed set-level concept formulation rather than a particular segmentation backend.

\subsection{Robustness to Judge Models}
\label{sec:judge_ablation}
To examine whether the annotation-quality results in Tab.~\ref{tab:ablation_annotation} are affected by same-family evaluation bias, we randomly sample 500 items from the MUSE test set and repeat the evaluation under the same protocol using Qwen3-VL-8B-Instruct, GPT-5.5, Gemini-3.1-Flash-Lite, Claude-Sonnet-4.6, and Doubao-Seed-2.0-lite as judges. As shown in Tab.~\ref{tab:judge_ablation}, diverse labeling consistently improves semantic-label quality across all five model families. Hierarchical semantics provides a further gain for every judge, confirming that the trend is robust to the choice of evaluation model.

\section{Efficiency Analysis}
\label{sec:efficiency}

\begin{table}[t]
    \centering
    \begin{minipage}[t]{0.58\textwidth}
        \centering
        \caption{\textbf{Image inference efficiency.} \modelname achieves substantially lower latency and computational cost than the compared methods.}
        \label{tab:efficiency_image}
        \vspace{2pt}
        \resizebox{\linewidth}{!}{%
            \begin{tabular}{lccrc}
                \toprule
                \textbf{Method} & \textbf{Params} & \textbf{Memory (GiB)} & \textbf{Latency (ms)} & \textbf{FLOPs (TFLOPs)} \\
                \midrule
                PixelLM & 7.15B & 13.7 & 1,877 & 9.9 \\
                SAM 3 Agent & 9.62B & 23.4 & 17,070 & -- \\
                VisionReasoner & 8.29B & 16.9 & 3,349 & 20.0 \\
                \rowcolor{gray!15}
                \textbf{\modelname (ours)} & \textbf{9.62B} & \textbf{18.6} & \textbf{878} & \textbf{5.5} \\
                \bottomrule
            \end{tabular}
        }
    \end{minipage}\hfill
    \begin{minipage}[t]{0.40\textwidth}
        \centering
        \caption{\textbf{Video inference efficiency.} \modelname achieves competitive throughput close to native SAM~3 inference.}
        \label{tab:efficiency_video}
        \resizebox{\linewidth}{!}{%
            \begin{tabular}{lccc}
                \toprule
                \textbf{Method} & \textbf{Params} & \textbf{Memory (GiB)} & \textbf{FPS} \\
                \midrule
                SAM 3 & 860M & 11.6 & 11.65 \\
                GLUS & 7.29B & 18.8 & 9.22 \\
                VISA & 7.70B & 28.3 & 4.63 \\
                \rowcolor{gray!15}
                \textbf{\modelname (ours)} & \textbf{9.62B} & \textbf{22.1} & \textbf{10.16} \\
                \bottomrule
            \end{tabular}
        }
    \end{minipage}
\end{table}

\begin{wrapfigure}{r}{0.54\textwidth}
    \vspace{-15pt}
    \centering
    \includegraphics[width=\linewidth]{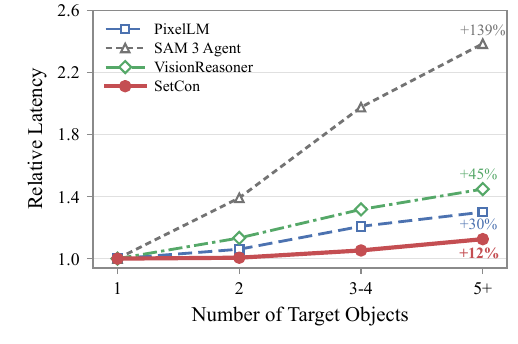}
    \captionsetup{width=\linewidth,justification=raggedright,singlelinecheck=false}
    \vspace{-20pt}
    \caption{\textbf{Latency scaling with the number of target objects.} \modelname scales favorably as the number of target objects increases, indicating efficient inference in complex multi-object scenarios.}
    \label{fig:efficiency_scaling}
    \vspace{-12pt}
\end{wrapfigure}

We conduct an efficiency analysis of \modelname against representative segmentation methods for both image and video tasks. All experiments are performed using one NVIDIA H200 GPU, 14 CPU cores, and 208~GiB of RAM. As reported in Tab.~\ref{tab:efficiency_image} and Tab.~\ref{tab:efficiency_video}, \modelname achieves significantly lower latency and computational cost for image inference, while maintaining competitive throughput for video inference.

We further investigate the latency scaling behavior with respect to the number of target objects. As shown in Fig.~\ref{fig:efficiency_scaling}, \modelname increases latency by only 12\% when scaling from one object to five or more objects, compared with 30\%, 45\%, and 139\% for PixelLM, VisionReasoner, and SAM~3 Agent, respectively. This favorable scaling is attributed to \modelname's unified prediction interface, which avoids object-by-object inference and enables efficient multi-object segmentation.

\section{Limitations}
Despite its promising results, our work still leaves room for improvement. First, open-ended queries can have ambiguous target boundaries; explicit concepts make the model's assumptions interpretable, and future work could incorporate interactive clarification to resolve such ambiguity. Second, our hierarchical annotations, built with model-assisted pipelines and quality control, may still contain noise and have limited coverage of rare long-tail cases. More diverse data sources and finer-grained human validation could further improve robustness and open-world generalization.

\section{Broader Impact}
\label{sec:appendix_impact}

\modelname improves fine-grained visual perception, with potential applications in accessibility, image and video editing, AR/VR interaction, and embodied perception. Like other vision-language systems, however, its deployment in surveillance, biometric identification, or other high-stakes settings raises concerns around privacy, consent, and fairness. The explicit concept interface introduced in this work makes grounding decisions more interpretable and easier to audit than implicit-token alternatives, but it does not eliminate biases inherited from large-scale pretraining or source segmentation datasets, which are largely based on web imagery and may underrepresent certain regions, demographics, and contexts. We encourage the responsible use of our dataset and method, and explicitly discourage any applications that may infringe upon personal privacy or be deployed for harmful purposes.

All training and evaluation datasets used in this work are publicly released for academic research. We use them strictly under their original licenses, and our hierarchical semantic annotations are derived only from publicly available samples. All annotations and experimental results were generated solely for research purposes and follow ethical guidelines for the use of public data in academic research.





\end{document}